\pdfoutput=1

\documentclass[11pt]{article}

\usepackage{acl}
\usepackage{times}
\usepackage{latexsym}

\usepackage[T1]{fontenc}

\usepackage[utf8]{inputenc}

\usepackage{microtype}
\usepackage{amsthm}

\usepackage{rotating,tabularx,siunitx}
\usepackage{comment}
\usepackage{float}
\usepackage{graphicx}
\usepackage{subfigure}
\usepackage{caption}
\usepackage{amsmath,amssymb}
\usepackage{mathtools}
\usepackage{booktabs}
\usepackage{multirow}
\usepackage{siunitx}
\usepackage{adjustbox}
\usepackage{pifont}
\usepackage{scalerel}
\usepackage{tabularray}
\usepackage{makecell}
\usepackage[misc,geometry]{ifsym} 

\newtheorem{definition}{Definition}
\usepackage{inconsolata}

\title{PrivLM-Bench: A Multi-level Privacy Evaluation Benchmark for Language Models}

\author{{\bf Haoran Li\footnotemark[1]}\textsuperscript{ \ }$^1$, {\bf Dadi Guo\footnotemark[1]}\textsuperscript{ \ }$ ^2$, {\bf Donghao Li\footnotemark[1]}\textsuperscript{ \ }$^{1}$, {\bf Wei Fan}$^{1}$, {\bf Qi Hu}$^{1}$\\
{\bf Xin Liu}$^{3}$, {\bf Chunkit Chan}$^{1}$,  {\bf Duanyi Yao}$^{1}$, {\bf Yuan Yao}$^{1}$, {\bf Yangqiu Song}$^{1}$\\
$^{1}$The Hong Kong University of Science and Technology\\
$^{2}$Center for Data Science, AAIS, Peking University 
$^{3}$Amazon.com Inc, Palo Alto, USA \\
\texttt{\{hlibt, dlibf, wfanag, qhuaf, ckchancc, dyao\}@connect.ust.hk},
 \\
 \texttt{guodadi@stu.pku.edu.cn},  \texttt{xliucr@amazon.com},
 \texttt{yuany@ust.hk}, \texttt{yqsong@cse.ust.hk}
\\
}

\begin{document}
\maketitle
{
\renewcommand{\thefootnote}{\fnsymbol{footnote}}
\footnotetext[1]{Equal contribution.}
}

\begin{abstract}
The rapid development of language models (LMs) brings unprecedented accessibility and usage for both models and users.
On the one hand,  powerful LMs achieve state-of-the-art performance over numerous downstream NLP tasks.
On the other hand, more and more attention is paid to unrestricted model accesses that may bring malicious privacy risks of data leakage.
To address these issues, many recent works propose privacy-preserving language models (PPLMs) with differential privacy (DP).
Unfortunately, different DP implementations make it challenging for a fair comparison among existing PPLMs. 
In this paper, we present PrivLM-Bench, a multi-perspective privacy evaluation benchmark to empirically and intuitively quantify the privacy leakage of LMs.
Instead of only reporting DP parameters, PrivLM-Bench sheds light on the neglected inference data privacy during actual usage.
PrivLM-Bench first clearly defines multi-faceted privacy objectives.
Then, PrivLM-Bench constructs a unified pipeline to perform private fine-tuning.
Lastly, PrivLM-Bench performs existing privacy attacks on LMs with pre-defined privacy objectives as the empirical evaluation results.
The empirical attack results are used to fairly and intuitively evaluate the privacy leakage of various PPLMs. 
We conduct extensive experiments on three datasets of GLUE for mainstream LMs.\footnote{Code is publicly available at \url{https://github.com/HKUST-KnowComp/PrivLM-Bench}.}
\end{abstract}

\section{Introduction}
The accelerating evolution of language models (LMs) ushers a new era for both modern natural language processing and the whole society.
Currently, generative large language models (LLMs) exhibit surprising capability and integrate previous tasks into a unified text generation formulation.
As a result, these LLMs obtain the dominating performance on both expert-designed tasks and real-world problems~\cite{2020t5,2022flant5,Brown2020LanguageMA,OpenAI2023GPT4TR, ouyang2022training}.
Moreover, under appropriate instructions, LLMs can even be in-context learners or zero-shot reasoners to solve unseen tasks~\cite{Chen2021EvaluatingLL, zhou2023leasttomost, Kojima2022LargeLM, Wei2022ChainOT,sanh2022multitask}.

Beneath the improved performance, LMs' training data also scale up with models' sizes.
LMs are not only trained on annotated textual data for specific tasks, but also devour a vast amount of textual data online.
Unlike carefully crowd-sourced annotation data, free-form texts crawled from the Internet suffer from poor quality and unintended personal data leakage.
For example, simple model interactions can lead to accidental personally identifiable information (PII) dissemination \cite{LI-2023-Jailbreak, Lukas2023AnalyzingLO, huang-etal-2022-large,carlini-2021-extracting}.
Such PII exposure without noticing victims or obtaining victims' consent may violate existing privacy laws like the EU's General Data Protection Regulation (GDPR) and the California Consumer Privacy Act (CCPA).

To respect data subjects' privacy during model training/fine-tuning, several studies~\cite{qu2021natural,yue2022synthetic,yu2022differentially,Igamberdiev-2023-DP-BART} consider privacy protection as an additional objective.
Differential privacy~\cite{Dwork-08-DP}, known for its wide adaptability and application, has become mainstream for privacy-preserving LMs (PPLMs).
DP's definition offers \textit{plausible deniability}~\cite{Bindschaedler-2017-Plausible} and introduces bounded privacy parameters to describe the effectiveness of examined mechanisms.
This definition can naturally defend against membership inference attack~\cite{Shokri-2016-MembershipIA}, which aims to determine if a given sample belongs to the model's training dataset.
Currently, motivated by DPSGD~\cite{Abadi2016DeepLW}, PPLMs can be achieved via various implementations based on DP optimizers.

Unfortunately, although numerous implementations of PPLMs have been proposed, fair evaluations of PPLMs are still unexplored.
Existing mainstream approaches simply use DP parameters to quantify PPLMs' privacy and it is rather hard to make a fair comparison among PPLMs.
Firstly, Different DP formulations such as central DP, local DP~\cite{kasiviswanathan-2011-localdp}, and $d_\chi$ privacy~\cite{Chatzikokolakis-2013-MetricDP} assign distinct heuristic meanings to DP parameters to implement PPLMs.
Secondly, the scope of the protected part is ambiguous.
For instance, most PPLMs implemented from DPSGD offer privacy protection for tuned sensitive data.
However, during inference, these PPLMs are not guaranteed to protect inference data privacy.
Thus, simply claiming that these PPLMs are privacy-preserving ignores inference data privacy.
Lastly, it remains unclear whether DP's worst-case upper bound overestimates privacy leakage.
DP assumes a herculean adversary who can manipulate the entire protected dataset, which may be implausible for actual attacks.
Consequently, evaluation under the same DP parameters may still result in different privacy performance on empirical privacy attacks.

To bridge the aforementioned gap, in this work, we propose PrivLM-Bench to fairly quantify PPLMs' privacy-utility trade-off.
PrivLM-Bench adopts the prevailing setup of public pre-training and private fine-tuning with several clarified privacy objectives.
PrivLM-Bench incorporates multi-faceted privacy attacks, including the data extraction attack~\cite{carlini-2021-extracting}, membership inference attack~\cite{Shokri-2016-MembershipIA} and embedding-level privacy attack~\cite{song-information-2020} to evaluate the privacy of PPLMs.
The attacking results can be an intuitive and fair indicator to quantify the privacy leakage of existing PPLMs regardless of their detailed implementations.
In summary, we highlight the following contributions of our proposed PrivLM-Bench:


1) PrivLM-Bench identifies inference data privacy as a key component for PPLMs' privacy evaluation and points out that DP-tuning cannot quantify the inference data privacy after deploying PPLMs for real-life applications.

2) PrivLM-Bench provides a unified pipeline that allows fair comparisons among PPLMs.

3) We use PrivLM-Bench to conduct extensive experiments on mainstream PPLMs. 
Empirical findings indicate that current privacy attacks are significantly less potent than the anticipated attacker capabilities of defense mechanisms.

\section{Related Works}
\subsection{Differential Privacy}
To analyze differential privacy implementations on language models, we first introduce the formal definition of DP~\cite{Dwork-08-DP}:

\begin{definition}[Differential Privacy]
\label{def:dp}
A randomized \textit{algorithm mechanism} $M$ with domain $D$ and range $R$ satisfies $(\epsilon,\delta)$-\textit{differential privacy} if for any two neighboring datasets $D, D'$ and for any subsets of output $O \subseteq R$:
\begin{equation}\label{eq:dp-bound}
Pr[M(D) \in O] \leq e^{\epsilon}Pr[M(D')\in O]+\delta.
\end{equation}
\end{definition}
The neighboring datasets $D, D'$ only differ in one element.
When we apply Definitions~\ref{def:dp} on LMs, $D$ refers to the private fine-tuning dataset and $M$ usually refers to the LM updated with DP mechanism so that the LM can be safely released while preserving $D$'s privacy.
For our experiments, DP optimizers, such as DPSGD~\cite{Abadi2016DeepLW}, are used as the backbone to implement various PPLMs with DP guarantee.

\subsection{Implementations on PPLMs}
There are several optional techniques to build PPLMs.
Homomorphic Encryption (HE) can ensure PPLM's data privacy via encryption during the inference stage~\cite{chen2022x}.
Secure Multiparty Computation (SMPC) protects the privacy of shared data and model parameters between service providers and users~\cite{wang2022characterization, hao2022iron, Luo-Practical-23, luo2024secformer}.
Existing works commonly exploit various DP mechanisms to implement DP-based LMs with respect to given fine-tuning corpus and can be summarized into 3 categories:
1): DP fine-tuning with DP optimizers~\cite{qu2021natural, shi-etal-2022-just,mattern-2022-differentially,yue2022synthetic,  li2022when,yu2022differentially} is a prevailing approach to protect the private fine-tuning datasets.
2): DP prompt tuning~\cite{duan2023flocks,li2023privacy} adds noise to soft prompts and performs private prompt tuning~\cite{lester-2021-prompttuning,li-2021-prefix} with LMs' parameters frozen.
3): Embedding perturbation~\cite{Igamberdiev-2023-DP-BART, Feyisetan-20-Privacy, krishna-2021-adept} injects DP noise into the intermediate representations to implement PPLMs.

For privacy evaluation, most of these works simply report $(\epsilon,\delta)$ pairs as the privacy budget.
Still, a few works endeavor to measure or explain privacy alternatively.
Auditing mechanisms~\cite{nasr2023tight,Jagielski-2020-Auditing,lu2022a} aim to audit empirical privacy leakage.
\citet{Feyisetan-20-Privacy} proposed plausible deniability statistics to quantify plausible deniability given DP parameters.
\citet{li2023privacy} and \citet{Du-Sanitizing-2023} evaluated privacy via empirical embedding-level attacks~\cite{song-information-2020}.
Motivated by such empirical evaluations, PrivLM-Bench further clarifies the privacy objectives and integrates more attacks for empirical privacy evaluation.

\begin{figure*}[t]
\centering
\includegraphics[width=0.95\textwidth]{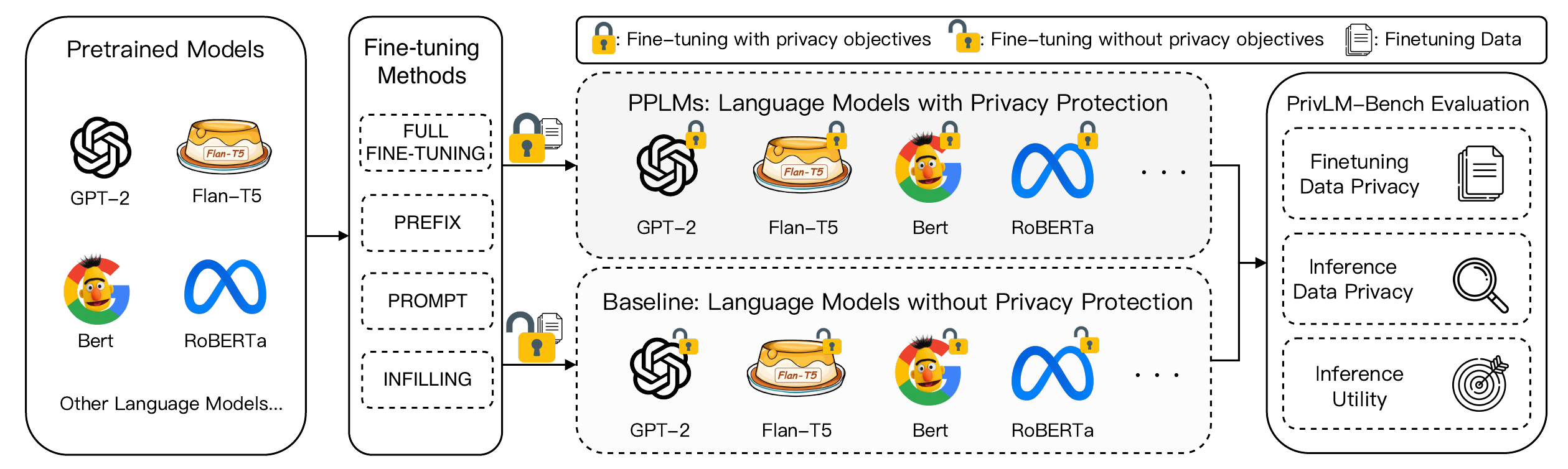}
\vspace{-0.1in}
\caption{
Overview of PrivLM-Bench.
}
\label{fig:PrivLM-Bench}
\vspace{-0.15in}
\end{figure*}

\section{PrivLM-Bench}

\subsection{Setup of Privacy Evaluation}
\label{sec:Evaluation Formulation}
PrivLM-Bench's scope follows the mainstream PPLMs' approaches that pre-train LMs on public corpus and then fine-tune these publicly pre-trained LMs on private datasets $D$ with privacy constraints.
Compared with publicly available pre-training data, private fine-tuning data $D$ are more likely to include or entail a set of sensitive attributes $P$.
For example, $x_s = $``Alice's phone number is +1 217-123-4567'' includes Alice's personal phone number and the area code ``217'' entails that Alice is likely to live in Illinois of the U.S.
Sensitive attributes $P$ may contain PII protected by privacy laws.
Thus, protection of $P$ should be carefully enforced.

With the pre-trained initialization and sensitive fine-tuning data $D$, we can implement and release a PPLM $f$.
Then, $f$ can be used on downstream tasks for inference without additional training/fine-tuning.
Similarly, we use $I$ to denote the inference data.
PrivLM-Bench manages to quantify $f$'s privacy regardless of its detailed implementation.

\subsection{What Should be Regarded as Privacy for PPLMs?}
To quantify a given LM $f$'s privacy-preserving ability, we first clarify PrivLM-Bench's privacy objectives. 
PrivLM-Bench includes multi-level privacy evaluation from fine-tuning to the inference stage.

\textbf{Privacy protection of fine-tuning data.}
To protect private fine-tuning data $D$, PrivLM-Bench considers two privacy objectives to address data leakage. 
The first objective focuses on protecting textual samples of $D$ while the second emphasizes its sensitive attributes.
Specifically, $f$ should prevent attackers from directly decoding the data sample $x \in D$ during the decoding stage.
Additionally, $f$'s hidden representations (e.g., logits and embeddings) should be less confident about memorizing sensitive patterns of $D$ when the adversary queries $f$ with input $x \in D$.

\textbf{Privacy protection during inference stage.}
DP tuning provides a theoretical privacy bound for protecting $D$ to safely release LMs.
However, the privacy of inference data $I$ is unprotected and unspecified by DP since the noise injected into the fine-tuning stage is irrelevant to $I$.
Consequently, PPLMs that merely report DP parameters neglect inference data privacy during practical model usage.
Additionally, from the data privacy perspective, inference data $I$ also inherently encompasses certain private information.
For instance, a healthcare facility may release its PPLM to serve patients.
If patients' unprotected medical records are recovered by malicious attackers, such data leakage may violate existing privacy legislation and frustrate its users.
Therefore, it is imperative to take the privacy of $I$ into consideration for $f$'s privacy evaluation.

\subsection{Evaluation via Privacy Attacks}
To measure privacy following predefined privacy objectives, we propose to evaluate PPLMs via a unified attack pipeline and use the performance of various attacks as privacy metrics.
PrivLM-Bench assumes a black-box adversary who can query $f$ with arbitrary inputs and obtain response texts as well as $f$'s hidden representations.
Moreover, the adversary has powerful knowledge about $D$ and $I$ and owns its auxiliary datasets $A_D$ and $A_I$.
$A_D$ and $A_I$ share similar distributions with $D$ and $I$, respectively, such that the adversary can use them to train powerful neural attackers.
Intuitively, better privacy protection is attained for any attack with worse attack performance.
Based on privacy objectives, the adversary's attacks are classified into three folds to evaluate any given $f$.

\begin{table*}[t]
\centering
\small
\begin{tabular}{l c c c c}
\hline
\textbf{Privacy Attacks} & \textbf{LM Vulnerability} & \textbf{Targeted Objective} & \textbf{Adv Cap} & \textbf{Metrics}\\ 
\hline
Data extraction attacks (DEAs)        & Text completion           & Fine-tuning data privacy  & $f$, prefixes & Exposure      \\
Membership inference attacks (MIAs)    & Predicted likelihood      & Fine-tuning data privacy  & $f, f(x), A_D$ & AUC, TPR      \\
Embedding inversion attacks (EIAs)     & Hidden representation     & Inference data privacy  & $f, f(x), A_I$   & micro-F1    \\
\hline
\end{tabular}
\vspace{-0.05in}
\caption{\label{tab:atk_intro}
A summary for evaluated privacy attacks, where Adv Cap refers to the adversary capability of assessed information, AUC stands for the area under the ROC curve, and TPR denotes the true positive rate.
}
\vspace{-0.15in}
\end{table*}

\textbf{Data extraction attacks.}
Data extraction attacks (DEAs)~\cite{carlini-2021-extracting} assume that the adversary knows certain prefixes which refer to partial textual patterns of $D$ or $S$.
For a textual sample $x = [p || s] \in D$ where x is split into a length-$k$ prefix $p$ and a suffix $s$, DEAs aim to recover the corresponding suffix $s$ via prompting $f$ with given prefix $p$.
$p$ can be an empty string if the adversary has no knowledge about $D$.
To conduct DEAs on generative LMs, we insert several pre-defined \textit{canary}~\cite{Carlini-2019-SS} patterns with randomly generated fake sensitive information into the training dataset. 
We assign different repetition times for each canary and report the exposure~\cite{Carlini-2019-SS} metric.
The exposure metric uses the rank of a specific pattern over potential candidates by comparing LMs' perplexity (confidence).
Memorized patterns frequently lead to higher model confidence with lower perplexity, resulting in higher exposure.

\textbf{Membership inference attacks.}
Membership inference attacks (MIAs)~\cite{Shokri-2016-MembershipIA} assume that the adversary holds its auxiliary data $A_D$ such that $A_D$'s partial samples are also members of $f$'s fine-tuning corpus $D$.
MIAs manage to determine if a given data sample $x \in A_D$ belongs to $D$.
We follow Likelihood Ratio Attacks (LiRA)~\cite{Carlini2021MembershipIA} to conduct MIAs on victim LMs by training multiple shallow models.
Area under the ROC Curve (AUC) scores and true positive rate (TPR) under certain fixed false positive rates are reported as MIA evaluation metrics.

\textbf{Embedding-level privacy attacks.}
Understanding the privacy of vector data in vector databases~\cite{taipalus2023vector, wang2021milvus,pan2023survey} is crucial, and embedding-based attacks play a fundamental role in this context.
These attacks
~\cite{song-information-2020,Pan-2020-Privacy,li-2022-dont} assume that the adversary possesses its auxiliary data $A_I$ that shares a similar distribution with private data $D$ and can access $f$'s embeddings during the inference stage.
Embedding-level attacks encompass attribute inference attacks (AIAs) and embedding inversion attacks (EIAs).
For a data sample $x \in I$, AIAs aim to infer private attributes of $x$ given its embedding $f(x)$ while EIAs focus on recovering $x$ given $f(x)$.
To study the inference stage privacy leakage, we follow the recently proposed generative EIAs~\cite{li-etal-2023-sentence,morris2023text,Gu2023TowardsSL} to use powerful decoders to recover the exact sequences and report micro-level precision, recall and F1 as the evaluation metrics.

In summary, Table~\ref{tab:atk_intro} lists the covered attacks for PrivLM-Bench's privacy evaluation. Full attack details can be found in Appendix~\ref{app: atk}.

\subsection{Potential Applications}
Based on the formulated pipeline, the PrivLM-Bench can be used for three potential applications.

\textbf{Comparison among PPLMs.}
PrivLM-Bench enables a fair comparison among various PPLMs of different architectures, learning objectives and pre-training data to conduct private fine-tuning.
This comparison helps service providers select the most suitable PPLM for target downstream tasks.

\textbf{Develop new attacks and defenses.}
PrivLM-Bench incorporates several existing privacy attacks into a unified pipeline. 
New attacks and defenses can be easily developed into PrivLM-Bench to conduct red-teaming assessments on LMs and PPLMs to evaluate their effectiveness.

\textbf{PPLM implementations' verification.}
Most PPLMs report DP parameters for privacy evaluation.
However, it remains unknown if their implementations are correct.
PrivLM-Bench provides an empirical privacy evaluation to verify the correctness of these implementations.

\section{Experimental Setups}
\subsection{Datasets}
Existing PPLMs evaluate their claimed improvement over tailored downstream tasks.
These specific tasks may not be feasible for other PPLMs.
Instead, PrivLM-Bench evaluates PPLMs in a more fundamental aspect for natural language understanding (NLU).
NLU is essential for general LMs to identify the meaning of given texts.
PrivLM-Bench evaluates PPLMs on several NLU tasks from GLUE ~\cite{wang-etal-2018-glue} that include Natural Language Inference (MNLI) \cite{Williams-mnli-2018}, Stanford Sentiment Treebank v2 (SST2) \cite{socher-etal-2013-recursive} and QNLI converted from Stanford Question Answering Dataset ~\cite{rajpurkar-etal-2016-squad}.

\subsection{Data Pre-processing}
PrivLM-Bench's evaluated datasets, MNLI, SST2 and QNLI, can be naturally formulated as classification tasks with given labels.
However, besides BERT-style masked LMs, PPLMs also include generative LMs such as GPT-2 that behave poorly for conventional classification pipelines.
We additionally transform evaluated datasets to fit the generation pipeline.
Inspired by T5's formulation~\cite{2020t5}, given the premises and hypotheses of NLI datasets or sentences from SST2 with corresponding integer labels, we manually create textual templates with labels and concatenate prefix sentences with templates.
For example, given a sample includes \textit{premise}, \textit{hypothesis} with label 0,  the converted sample for generation becomes a single sentence [\textit{premise} $<$SEP$>$ \textit{hypothesis} $<$SEP$>$  The relation is 0] where $<$SEP$>$ is the special separator token.
We keep the integer label in the transformed sentence for easier text-infilling formulation.
After the pre-processing, PrivLM-Bench can evaluate most existing PPLMs.

\subsection{Pretrained Weights and Fine-tuning Methods}
Since PrivLM-Bench unifies both classification and generation tasks, various model architectures with different pre-trained weights can be evaluated by PrivLM-Bench.
Specifically, We evaluate BERT~\cite{devlin-etal-2019-bert}, RoBERTa~\cite{Liu2019RoBERTaAR}, GPT-2~\cite{radford-2019-language}, T5~\cite{2020t5} and FLAN-T5~\cite{flant5-2022} 
of different scales with four tuning algorithms with and without the DP guarantee. 
The following content gives a brief summary of our evaluated tuning algorithms.

\textbf{Full fine-tuning.}
Full fine-tuning refers to the commonly used methods to update the whole model. 
For masked LMs such as BERT and RoBERTa, we append one extra linear layer to perform classification.
For generative LMs, including GPT-2 and T5, we use language modeling head with language modeling loss to conduct the next token generation.

\textbf{Prompt tuning}~\cite{lester-2021-prompttuning}.
For prompt tuning, we freeze the LMs and prepend certain learnable virtual tokens to every data sample.
Instead of updating the whole LMs, prompt tuning only optimizes these task-specific virtual tokens.

\textbf{Prefix tuning}~\cite{li-2021-prefix}.
Prefix tuning shares a similar idea as prompt tuning.
Unlike prompt tuning which appends a few tokens in the beginning, prefix tuning attaches a sequence of continuous task-specific vectors in front of inputs.
These appended continuous vectors are concatenated into LMs' hidden states in every transformer layer as well as inputs.
Only these appended vectors are updated throughout prefix tuning.

\textbf{Infilling based tuning}~\cite{petroni-etal-2019-language}.
For masked LMs, instead of appending classifiers to their final output representations, infilling the masked tokens can also be exploited to perform classification tasks.
We follow the pre-processing pipeline to leave blanks for predicted positions, like
[\textit{premise} $<$SEP$>$ \textit{hypothesis} $<$SEP$>$  The relation is $<$MASK$>$], where $<$MASK$>$ refers to the mask token.
For infilling-based tuning, we follow previous works to update the whole LMs.

\begin{table*}[t]
\centering
\small
\fontsize{9pt}{9pt}\selectfont
  \begin{tabular}{l| cccc| cccc}
    \toprule
    \multirow{2}{*}{Model} &

      \multicolumn{4}{c|}{Non-DP $(\epsilon= \infty)$} &
      \multicolumn{4}{c}{DP $(\epsilon=\text{8},\delta=\text{1e-5})$} 
       \\

      {} & {Fine-tuning} & {Prompt} & {Prefix} & {Infilling} & {Fine-tuning} & {Prompt} & {Prefix} & {Infilling} \\

    \midrule

    BERT\textsubscript{base} (110M)
    & 82.97 & 66.38 & 79.20 & 82.05 & 65.19 & 47.44 & 43.69 & 61.57  \\
    BERT\textsubscript{large} (340M)
    & 85.31 & 66.17 & 83.63 & \textbf{84.70} & 71.90 & \textbf{61.67} & 52.30 & \textbf{63.41}  \\
    RoBERTa\textsubscript{base} (125M)
    & 87.45 & 67.65 & 83.97 & - & 76.45 & 46.26 & 49.16 & -  \\
    RoBERTa\textsubscript{large} (355M)
    & 90.10 & \textbf{70.49} & \textbf{89.74} & - & \textbf{83.53} & 60.81 & 64.10 & -  \\
    GPT-2\textsubscript{small} (137M)
    & 39.29 & 37.02 & 34.25 & - & 34.96 & 35.81 & 36.85 & -  \\
    GPT-2\textsubscript{medium} (380M)
    & 63.37 & 35.01 & 43.86 & - & 41.31 & 33.25 & 35.17 & -  \\
    GPT-2\textsubscript{large} (812M)
    & 76.48 & 31.28 & 52.44 & - & 46.75 & 33.28 & 35.05 & -  \\
    GPT-2\textsubscript{xl} (1.6B)
    & 82.88 & 34.37 & 44.37 & - & 47.73 & 32.37 & 35.17 & -  \\
    T5\textsubscript{base} (223M)
    & 86.33 & 32.23 & 75.62 & - & 78.50 & 31.64 & 64.87  & -  \\
    T5\textsubscript{large} (738M)
    & 89.13 & 35.59 & 72.69 & - & 54.85 & 31.67 & 34.79 & -  \\
    T5\textsubscript{xl} (3B)
    & 91.10 & 33.18 & 83.32 & - & OOM & 38.63 & 66.51 & -  \\
    FLAN-T5\textsubscript{xl} (3B)
    & \textbf{92.07} & 34.79 & 87.94 & - & OOM & 33.45 & \textbf{83.54} & -  \\
    \bottomrule
  \end{tabular}
  \vspace{-0.05in}
  \caption{\label{tab:utility}
Utility evaluation of various LMs with and without DP tuning on the MNLI dataset. We report the accuracy (\%) of the validation set as the models' utility. OOM refers to out of memory for our GPU devices.
}
\vspace{-0.1in}
\end{table*}
\subsection{Other Details}

\textbf{Data split.} 
Both membership inference attacks and embedding-level attacks require auxiliary datasets that share similar distributions with original datasets. 
For each evaluated dataset, we randomly split 40\% training data as the auxiliary dataset and use the remaining 60\% to conduct downstream tuning.

\textbf{DP Parameters.} During our experiments, we follow previous studies~\cite{li2022large,YuZC0L21} to strictly bound $\delta$ = 1e-5 and $\epsilon$ = 8.
Additionally, we bound the gradient norms to be no more than 0.1.

\textbf{Tuning parameters.}
For all four tuning algorithms, we use adamW as the optimizer with a linear decay.
We train all models for 5 epochs with a virtual batch size of 1,024.
In terms of learning rates, for fine-tuning and infilling, the learning rate is 1e-4; for prompt and prefix tuning, the learning rate is 1e-2.
We set 15 virtual tokens for optimization for both prompt and prefix tuning.


\section{Experiments}
For experiments, we first raise a few crucial research questions (RQs) and use our experimental results to address these RQs individually.

$\bullet$ \textbf{RQ1}: Do LMs share similar utility under the same DP budget? If not, what are the factors that affect LMs' utility?

$\bullet$ \textbf{RQ2}: Do various tuning algorithms yield similar performance on the same model?

$\bullet$ \textbf{RQ3}: Are empirical privacy attacks effective on LMs with and without privacy protection?



\begin{table*}[t]
\centering
\small
\fontsize{9pt}{9pt}\selectfont
\setlength\extrarowheight{1pt}
  \begin{tabular}{lclc|  ccc| ccc}
    \toprule
    \multirow{2}{*}{Model} &
    \multirow{2}{*}{DP?} &
    \multirow{2}{*}{Tuning} &
    \multirow{2}{*}{Utility} &
      \multicolumn{3}{c|}{MIA} &
      \multicolumn{3}{c}{EIA} 
       \\
      {} & {} & {} & {}  & {AUC} & TPR@0.1\% & TPR@1\%  & Pre & Rec & F1 \\
      \midrule

      \multirow{8}{*}{BERT\textsubscript{base}} 
    & \multirow{3}{*}{Y}         
    & Finetune 
    & 77.72 & 50.12 & 0.11 & 0.98 & 44.22 & 18.05 & 25.63 \\
    &     & Prompt
    & 71.74 & 50.06 & 0.25 & 0.94 & 42.44 & 18.20 & 25.47 \\
    &     & Prefix
    & 68.29 & 49.96 & 0.12 & 1.08 & 44.62 & 18.41 & 26.06 \\

    \cmidrule{2-10}
    
   & \multirow{3}{*}{N}         
    & Finetune 
    & 90.14 & 50.14 & 0.09 & 1.05 & 41.57 & 17.71 & 24.84\\
    &     & Prompt
    & 79.34 & 51.40 & 0.15 & 1.36 & 43.80 & 18.53 & 26.05\\
    &     & Prefix
    & 87.45 & 51.19 & 0.16 & 1.38 & 44.17 & 18.41 & 25.98\\
    
    \midrule

    \multirow{8}{*}{BERT\textsubscript{large}} 
    & \multirow{3}{*}{Y}         
    & Finetune 
    & 81.14 & 49.80 & 0.12 & 0.97 & 42.29 & 17.65 & 24.91 \\
    &     & Prompt
    & 71.55 & 50.03 & 0.19 & 1.05 & 40.67 & 17.55 & 24.52 \\
    &     & Prefix
    & 69.05 & 49.77 & 0.12 & 1.00 & 42.68 & 17.92 & 25.24 \\

    \cmidrule{2-10}
    
   & \multirow{3}{*}{N}         
    & Finetune 
    & 90.93 & 50.02 & 0.09 & 1.01 & 41.38 & 17.30 & 24.40 \\
    &     & Prompt
    & 76.07 & 49.91 & 0.08 & 0.91 & 41.97 & 17.89 & 25.09 \\
    &     & Prefix
    & 87.94 & 50.58 & 0.14 & 1.25 & 43.08 & 17.39 & 24.77 \\
    


    \midrule
    \multirow{6}{*}{RoBERTa\textsubscript{large}} 
    & \multirow{3}{*}{Y}         
    & Finetune 
    & 84.31 & 50.20 & 0.11 & 1.02 & 37.96 & 15.62 & 22.13 \\
    &     & Prompt
    & 73.32 & 50.55 & 1.16 & 1.16 & 39.47 & 16.40 & 23.17 \\
    &     & Prefix
    & 72.71 & 49.74 & 0.62 & 1.06 & 38.12 & 16.15 & 22.69 \\
    \cmidrule{2-10}
    & \multirow{3}{*}{N}         
    & Finetune 
    & 93.43 & 61.12 & 3.33 & 9.25 & 34.17 & 14.66 & 20.52 \\
    &     & Prompt
    & 81.62 & 49.83 & 0.08 & 1.01 & 39.45 & 16.40 & 23.17 \\
    &     & Prefix
    & 92.21 & 50.39 & 0.08 & 1.11 & 37.76 & 15.98 & 22.45 \\
    

    \midrule
    \multirow{6}{*}{GPT-2\textsubscript{medium}} 
    & \multirow{3}{*}{Y}         
    & Finetune 
    & 53.61 & 53.70 & 0.42 & 1.28 & 52.87 & 20.25 & 29.29 \\
    &     & Prompt
    & 51.84 & 50.02 & 0.09 & 0.82 & 52.85 & 20.53 & 29.57 \\
    &     & Prefix
    & 48.51 & 49.86 & 0.12 & 1.06 & 55.84 & 21.50 & 31.04 \\
    \cmidrule(lr){2-10}    
   & \multirow{3}{*}{N}         
    & Finetune 
    & 58.22 & 59.28 & 0.67 & 2.31  & 56.64 & 21.80 & 31.48 \\
    &     & Prompt
    & 52.63 & 50.02 & 0.18 & 0.76  & 53.96 & 20.77 & 29.99 \\
    &     & Prefix
    & 51.02 & 49.87 & 0.27 & 1.15  & 51.08 & 20.47 & 29.22 \\


    \midrule
    \multirow{6}{*}{GPT-2\textsubscript{xl}} 
    & \multirow{3}{*}{Y}         
    & Finetune 
    & 55.59 & 53.26 & 0.24 & 1.46 & 59.55 & 23.09 & 33.28 \\
    &     & Prompt
    & 50.93 & 49.28 & 0.03 & 0.54 & 58.24 & 22.78 & 32.75 \\
    &     & Prefix
    & 49.67 & 49.52 & 0.09 & 0.85 & 51.08 & 20.47 & 29.22 \\
    \cmidrule(lr){2-10}    
   & \multirow{3}{*}{N}         
    & Finetune 
    & 83.03 & 82.35 & 8.02 & 22.71 & 59.13 & 22.80 & 32.91 \\
    &     & Prompt
    & 51.93 & 48.73 & 0.06 & 0.88 & 58.40 & 22.87 & 32.87 \\
    &     & Prefix
    & 52.33 & 49.88 & 0.18 & 1.00 & 59.02 & 23.15 & 33.25 \\

    \bottomrule
  \end{tabular}
\vspace{-0.05in}
\caption{\label{tab:privacy_evaluation}
A complete privacy evaluation of the QNLI dataset. 
MIA and EIA results are reported in \%.
TPR@0.1\% and TPR@1\% denote true positive rate with fixed 0.1\% and 1\% false positive rate, respectively.
}

\vspace{-0.15in}
\end{table*}





\subsection{Utility Evaluation for RQ1 and RQ2}
By fixing values of $(\epsilon,\delta)$ pairs during DP tuning, we can compare PPLMs' utility between masked LMs and generative LMs.
In Table~\ref{tab:utility}, we comprehensively list various LMs' utility of different scales on the MNLI dataset.
These results suggest that LMs' utility is affected by multiple factors, including model architectures, pre-trained weights, model sizes, tuning algorithms and DP constraints.

\textbf{Model architectures with pre-trained weights}. By fixing a similar model size, for masked LMs, RoBERTa models outperform BERT models with around 10\% improved accuracy on fine-tuning and prefix-tuning and share comparable accuracy with BERT models on prompt-tuning.
In addition, under the same T5 model architecture of a similar size, FLAN-T5\textsubscript{xl} still significantly surpasses T5\textsubscript{xl} on both DP and non-DP settings.
Such utility improvement is likely to come from FLAN-T5's large-scale instruction tuning.
Moreover, for generative LMs, T5\textsubscript{base} even significantly surpasses GPT-2 models from the small size to the xl size in terms of fine-tuning and prefix tuning.
These results suggest that model architectures, including their pre-trained weights, play a crucial role in downstream tasks' utility for tuning with and without DP constraints.

\textbf{Model size}. 
For the same model, after observing the accuracy of different model scales of BERT, RoBERTa, GPT-2 and T5, we can see that increased model sizes are likely to bring better utility for both DP and non-DP tuning.

\textbf{DP constraints}. 
Under the exact tuning method of the same model, we can still observe that DP tuning leads to non-negligible utility degradation from fine-tuning to infilling.


\begin{figure*}[t]
\centering
\setlength{\abovecaptionskip}{-0.0cm} 
\subfigure[DEAs on GPT-2\textsubscript{large}.]{
\begin{minipage}[t]{0.48\textwidth}
\centering\vspace{-0.05in}
\includegraphics[width=\linewidth]{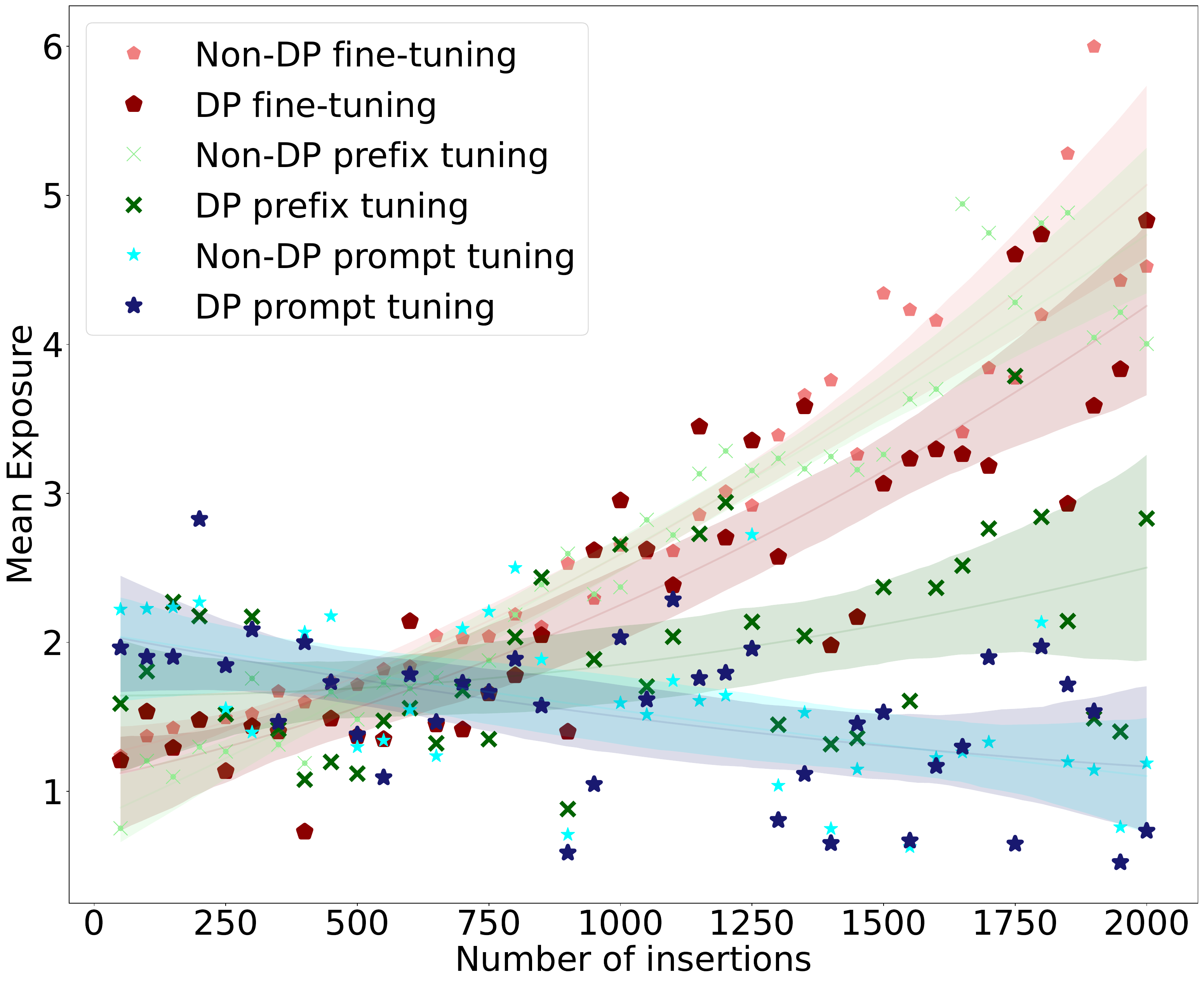}
\end{minipage}
}
\subfigure[DEAs on T5\textsubscript{large}.]{
\begin{minipage}[t]{0.48\textwidth}
\centering\vspace{-0.05in}
\includegraphics[width=\linewidth]{figs/exposure-T5-large.pdf}
\end{minipage}
}
\caption{DEA evaluation
 results of generative LMs on the MNLI dataset.}\label{fig:DEA}
\vspace{-0.15in}
\end{figure*}
\textbf{Tuning algorithms}. Additionally, results from Table~\ref{tab:utility} suggest that distinct LMs have varied resilience when we switch from non-DP tuning to DP tuning. For instance, for a given LM, fine-tuning mostly yields the best results with and without DP tuning and hurts \textasciitilde10\% accuracy for DP tuning.
However, when it comes to prompt and prefix tuning for masked LMs, severe accuracy drops can be observed.
Moreover, prompt tuning and prefix tuning are unreliable even for non-DP tuning for generative LMs such as GPT-2 models.
These results indicate that prefix tuning is better than prompt tuning for both DP and non-DP tuning under the same virtual token numbers.
For infilling-based knowledge probing, our results suggest that even though infilling has comparable performance with fine-tuning on the non-DP setting, infilling on noisily updates may suffer poorer results than direct fine-tuning for DP tuning.

\subsection{Privacy Evaluation for RQ3}
In this section, we perform privacy evaluations of the mainstream LMs with and without DP tuning. 
In Table~\ref{tab:privacy_evaluation}, we list masked LMs' privacy evaluation with empirical privacy attacks for both fine-tuning data privacy (MIAs) and inference data privacy (EIAs).
For generative LMs, we report the mean exposure of GPT-2\textsubscript{large} and T5\textsubscript{large} in Figure~\ref{fig:DEA}.

\textbf{MIAs}. 
For MIAs, all DP-tuned LMs achieve AUC scores around 0.5, indicating that MIAs on these models are no better than random guessing.
Hence, DP-tuned LMs conform to DP's definition and offer robust fine-tuning data protection against MIAs.
Conversely, for fine-tuned LMs without DP, MIAs are effective on RoBERTa models with AUC exceeding 0.6, showing that fine-tuned RoBERTa models are susceptible to MIAs.
In contrast, MIAs on BERT models without DP only gain marginal improvement of no more than 1.4\% compared to DP-tuned BERT models across three tuning algorithms.
These findings underscore a critical gap in the field of privacy attacks. 
The actual performance of attacks falls short of the level of threat assumed by existing defense mechanisms.

\textbf{EIAs}. 
For EIAs, since fine-tuning data privacy is not guaranteed by DP-tuning, we study the inference data privacy with and without DP tuning.
After comparing EIAs' performance differences between DP and non-DP settings among all models in Table~\ref{tab:privacy_evaluation}, we can observe the minimal variation with no more than 2\% deviation.
This observation indicates that DP tuning does not significantly alter the level of inference data leakage in EIAs compared to non-DP settings.
Thus, naive DP tuning on sensitive data cannot protect inference data privacy and requires additional privacy mechanisms to support the general privacy protection claim.

\textbf{DEAs}. 
For DEAs, we utilize canaries' exposure to measure generative LMs' training data privacy by increasing occurrences of certain canaries.
As shown in the scatter plots of Figure~\ref{fig:DEA}, the results reveal that DP-tuning is effective against DEAs with reduced exposure in comparison with corresponding non-DP tuning.
For DP-based tuning algorithms, even with the increased insertion numbers, DP-based prompt tuning and prefix tuning are effective against DEAs with low exposure.
This underscores the effectiveness of DP-based methods in mitigating DEAs.
However, when occurrences of given canaries increase, DP fine-tuning still suffers from non-neglected increased exposure, which requires larger noise to mitigate the exposure.
On the other hand, in terms of non-DP baselines, fine-tuning and prefix tuning suffer from high exposure when the insertion number increases, which conforms to previous works' findings on fine-tuning~\cite{carlini-2021-extracting}.
Unexpectedly,  our results show that prompt tuning can achieve relatively low exposure with and without DP.
Non-DP prompt tuning yields lower exposure even than DP-based prefix tuning and fine-tuning.

Upon analyzing the results, we summarize the following findings.
1) DP-tuning is robust against existing privacy attacks targeted at fine-tuning data. 
Regardless of tuning methods, DP's worst-case bounding can strictly prevent the adversary from identifying sensitive fine-tuning data.
2) Naive DP-tuning falls short in safeguarding inference data privacy. 
Our analysis reveals that DP-based methods do not significantly outperform non-DP baselines in preventing EIAs.
3) 
The capability of existing privacy attacks does not align with the level of threat presumed by defense strategies.
Notably, several non-DP tuning methods effectively resisted the evaluated attacks.
For MIAs, we show that LiRA is unsuccessful on non-DP BERT models.

\subsection{Other Empirical Findings}
In this section, we list other empirical findings according to the evaluation results.

1) Large generative LMs have the potential to outperform masked LMs with DP on NLU.
Our analysis, as presented in Table~\ref{tab:utility}, reveals that FLAN-T5\textsubscript{xl} with DP-based prefix tuning, can match or even surpass the performance of masked LMs with DP in NLU tasks.

2) Vulnerabilities to privacy attacks are model-dependent.
This is evident from the clustered patterns of attack performance for the same models of different sizes, as observed in Table~\ref{tab:privacy_evaluation} and Figure~\ref{fig:DEA}. 
This suggests that certain architectures may inherently possess more resilience to privacy breaches.

3) Parameter-efficient tuning methods are more resistant to privacy attacks.
In terms of attack results under non-DP settings, both prompt and prefix tuning exhibit greater robustness compared to full fine-tuning. This is particularly notable in their performance against MIAs and DEAs, indicating that these tuning methods inherently offer enhanced privacy safeguards.


\section{Conclusion}
In this paper, we introduce PrivLM-Bench, a benchmark designed to assess and contrast LMs' multifaceted privacy objectives.
By integrating a variety of masked and generative LMs with diverse tuning algorithms, PrivLM-Bench facilitates an extensive evaluation that encompasses both utility metrics and empirical privacy attacks.
Our results reveal the effectiveness and limitations of various DP tuning methods.
Moreover, we point out the reality that empirical privacy attacks demonstrate a less potent threat compared to defense mechanisms' assumed powerful capabilities.
In conclusion, our study underscores that privacy evaluation and the balancing act between privacy and utility is a complex, ongoing challenge in the field.
For future work, we advocate for more potent privacy attacks and utility-enhanced defense strategies that relax the worst-case restriction in accordance with empirical attacks to improve the privacy utility trade-off.

\newpage
\section*{Limitations}
In our evaluation of language model privacy from an adversarial standpoint, we acknowledge certain limitations in the covered scope and effectiveness of the proposed attacks. 
Firstly, our study does not encompass all attack methodologies, notably excluding the recent trend of prompt injection attacks, which are significant in assessing the safety of large language models. 
This omission represents an area for potential future exploration to provide a more comprehensive understanding of LLM vulnerabilities.
Secondly, the efficacy of our attacks in certain scenarios was limited. For instance, our LiRA-based MIAs' experiments demonstrated inadequate performance in discerning membership status in non-differentially private (non-DP) tuned BERT models. 

In addition, from defenders' perspectives, though DP tuned models can well protect LMs from inferring sensitive fine-tuning data, many DP tuned LMs suffer from poor utility.
Moreover, evaluated DP tuning strategies cannot defend against inference stage data privacy.

In summary, these limitations emphasize the need for ongoing development in the field of LM privacy attacks and defenses.

\section*{Ethical Considerations}
We declare that all authors of this paper acknowledge the \emph{ACM Code of Ethics} and honor the code of conduct.
This work comprehensively evaluates the empirical privacy of LMs and PPLMs via existing privacy attacks.
The purpose of these attacks is not to corrupt given LMs. Instead, we aim to test LMs' robustness against known attacks and show that DP can well protect the sensitive tuning data while fails to guarantee inference data privacy.
Our findings reveal that LLM still needs further improvement for the better privacy-utility trade-off.

\textbf{Data}. 
During our experiment, besides downstream fine-tuning data from GLUE tasks, we also pre-define several canary patterns with randomly generated or LLM-generated PII.
Since these canaries do not include any actual personal information, our inserted canaries can be safe to release and use.

\textbf{Attacks}. 
We are well aware that our evaluated privacy attacks may be used for malicious purposes. 
However, our experimental results that these empirical privacy attacks are rather weak in terms of privacy attacks and can be easily defended with defense mechanisms.
For example, data extraction attacks can be easily addressed with content filters to avoid unsafe responses.
We emphasize that these empirical privacy attacks are beneficial in enhancing defense strategies.

\section*{Acknowledgment}
The authors of this paper were supported in part by the NSFC Fund (U20B2053) from the NSFC of China, the RIF (R6020-19 and R6021-20), the GRF (16211520, 16205322, and 16308321) from RGC of Hong Kong, and the NSFC/RGC Joint Research Scheme Grant N\_HKUST635/20.
We also thank the support from the UGC Research Matching Grants (RMGS20EG01-D, RMGS20CR11, RMGS20CR12, RMGS20EG19, RMGS20EG21, RMGS23CR05, RMGS23EG08). 

\newpage
\bibliography{custom}

\appendix

\clearpage
\section{Training Details}
\label{app: exp_details}
During our experiment, we use 2 NVIDIA RTX 6000 to run our codes and it takes GPU hours around 2 months to complete all experiments.

\section{Attack Details}
\label{app: atk}
In this section, we explain the attack details used for our PrivLM-Bench including data extraction attack (DEA), membership inference attack (MIA) and generative embedding inversion attack (EIA).

\subsection{MIA Details}
We employ Likelihood Ratio Attacks (LiRA)~\cite{Carlini2021MembershipIA} to perform MIAs.

When attacking the encoder-only model, such as BERT and RoBERTa, we utilize the online version of LiRA. 
This version assumes that the attacker has obtained partial training data, which is used to construct shadow models.
We construct a total of 128 shadow models to estimate the logit distribution for each sample in the dataset. For each shadow model, the training set and test set are randomly divided.
Within the 128 shadow models, each sample is employed for both training and testing, enabling us to estimate the training logit distribution and testing logit distribution. We assume their distributions to be Gaussian and calculate the mean and variance for each sample.
Specifically, for a sample $x$ and shadow models $f_i$ where $i=1,2,...,128$, we can obtain its training logit distribution $\mathcal{N}(\mu_{\text{in}}, \sigma_{\text{in}}^2)$ and testing logit distribution $\mathcal{N}(\mu_{\text{out}}, \sigma_{\text{out}}^2)$.
Let $t(x,f_i):=\mathbf{1}\{f_i \text{ is trained on x}\}$ denote $f_i$ is trained on $x$, $f_i(x)$ denote the logits of the LM. Then, we can get the mean and variance of the logit distribution of $x$ as follows:

\begin{align}
\small
\alpha &= \frac{1}{\sum_{i=1 }^{128} t(x,f_i) },
\beta = \frac{1}{\sum_{i=1 }^{128} \neg t(x,f_i) }\\
\mu_{in}&=\alpha \sum_{i=1 }^{128} t(x,f_i)f_i(x)\\
\sigma_{in}^2&=\alpha \sum_{i=1 }^{128} t(x,f_i)(f_i(x)-\mu_{in})^2\\
\mu_{out}&=\beta \sum_{i=1 }^{128} \neg t(x,f_i)f_i(x)\\
\sigma_{out}^2&= \beta \sum_{i=1 }^{128} \neg t(x,f_i)(f_i(x)-\mu_{out})^2
\end{align}

After that, we can calculate the likelihood ratio of $x$ as follows:
$$\frac{p(x|\mu_{in}, \sigma_{in}^2)}{p(x|\mu_{out}, \sigma_{out}^2)}$$
The likelihood ratio is the ratio of the probability of $x$ in the train logit distribution and the probability of $x$ in the test logit distribution.
A larger likelihood ratio means the sample is more likely to be in the training set.

When attacking the decoder models, for example, GPT-2, we conduct a LiRA attack and choose the pre-trained model as the shadow model. 
Specifically, for a sample $x$, we calculate its likelihood from the target model and shadow model. We use the likelihood ratio as the indicator of membership. A larger likelihood ratio indicates that the sample is more likely to be in the training set. 


\begin{table*}
\small
\begin{center}
\begin{tabular}{l|p{3cm}|p{3cm}|c|p{3cm}}
\hline
\textbf{Type} & \textbf{Format for GPT-2} & \textbf{Format for T5} & \textbf{\# of Candidates} & \textbf{Notes} \\
\hline
Name & My name is $\bigcirc$ & Encoder: What is your name? Decoder: My name is $\bigcirc$ & 80 & - \\
\hline
City & I live in $\bigcirc$. It is a beautiful city & Encoder: Where do you live? Decoder: I live in $\bigcirc$. It is a beautiful city& 80 & - \\
\hline
Email & My email is $\bigcirc$ & Encoder: What is your email address? Decoder: My email is $\bigcirc$ & 70 & - \\
\hline
Phone Number & My phone number is $\bigcirc$. You can call me by this number. & Encoder: What is your phone number? Decoder: My phone number is $\bigcirc$. You can call me by this number. & 100 & The fill-in is composed of 5 random digits like "9 0 5 8 4". \\
\hline
Letters & The letters are $\bigcirc$. It is my password. & Encoder: What is your password? Decoder: The letters are $\bigcirc$. It is my password.& 100 & The fill-in is composed of 6 random letters from the alphabet like ' c z v s k o'. \\
\hline
One Word & [fixed word] [fixed word] $\bigcirc$ [fixed word] [fixed word] & Encoder: Choose one random word. Decoder: The chosen word is $\bigcirc$ & 100 & The canary is composed of four fixed words and \textbf{one} random word from the vocabulary of the tokenizer. \\
\hline
Three Words & [fixed word] [fixed word] $\bigcirc$ [fixed word] [fixed word] & Encoder: Choose three random words. Decoder: The chosen words are $\bigcirc$ & 100 & The canary is composed of four fixed words and \textbf{three} random words from the vocabulary of the tokenizer. \\
\hline
\end{tabular}
\vspace{-0.05in}
\caption{Different types of canary format and the number of full-in candidates of each canary format.}
\label{app:canary-format}
\vspace{-0.15in}
\end{center}
\end{table*}

\subsection{DEA Details}
For data extraction attacks on generative LMs, we insert different types of pre-defined \textit{canaries} \cite{Carlini-2019-SS} into the training dataset. Here we introduce the essential notations and the attack method, as well as some details in our experiments.

We begin with the definition of \textit{log-perplexity} which measures the likelihood of data sequences and the definition of \textit{canary}.

\begin{definition}[Log-Perplexity]
\label{log-ppl}
Given a model f with parameters $\theta$, the \textbf{log-perplexity} of a sequence $x_{1}...x_{n}$ is
\begin{align}
Px_{\theta}(x_{1}...x_{n}) &= -log_{2}\text{P}(x_{1}...x_{n}|f_{\theta})\\
&= \sum_{i=1}^{n} (-log_2\text{P}(x_i|f_{\theta}(x_{1}...x_{i-1})))\nonumber
\end{align}
\end{definition}

\textit{Canaries} refer to formatted sequences with some spaces infilled with random words or characters chosen from a specific \textit{randomness space $\mathcal R$}. In format sequences, the "slots" denoted as $\bigcirc$ can be filled with random values.  For the format "The random number is $\bigcirc \bigcirc \bigcirc \bigcirc \bigcirc$", each $\bigcirc$ might be filled with a specific random number if $\mathcal R$ refers to the numerical space of digits 0 to 9. We denote the randomness as \textit{r} and use the notation \textit{s}[\textit{r}] to indicate the format \textit{s} with holes filled with randomness \textit{r}.
Definitions of the \textit{rank} and \textit{exposure} of a canary are introduced below.

\begin{definition}
Given a model with parameters $\theta$ and the random space $\mathcal R$, the \textbf{rank} of a canary s[r] is
\begin{align}
\textbf{rank}_\theta({s}[{r}]) = |{r^{'} \in \mathcal R : Px_{\theta}({s}[r^{'}]) \leq Px_{\theta}({s}[{r}])}|
\end{align}
\end{definition}

The definition means the \textit{rank} of a specific canary is its index in the list of all possibly-instantiated canaries, ordered by the empirical model perplexity of all those sequences.

\begin{definition}[exposure]
Given a canary s[r], a model with parameters $\theta$ and the random space $\mathcal R$, the \textbf{exposure} of s[r] is
\begin{align}
\textbf{exposure}_{\theta}(s[r]) = log_2|\mathcal R| - log_2|\textbf{rank}_{\theta}(s[r])|
\end{align}
\end{definition}

There are some properties of the exposure. 1) It is a real value ranging between 0 and $log_2|\mathcal R|$. 2) Its maximum can be achieved only by the most likely, top-ranked canary while its minimum of 0 is the least likely. 3) Note that across possibly-inserted canaries, the median exposure is 1.

In our experiment, we prepare different canary types. 
For example, one format of our prepared canaries is "I live in $\bigcirc$. It's a beautiful city." while another format is  "My email is $\bigcirc$." We prepare multiple candidates for each format. 
For example, we use a list of city names like ["Tokyo, Japan", "London, United Kingdom", "Beijing, China", "New York, United States", ...] and fake email addresses generated by a large language model(not used in our experiments) like ["Liam@yahoo.com", "Emma@outlook.com", " Olivia@gmail.com", ...] for the above mentioned two formats. 
Meanwhile, we also follow the setting of the original \textit{secret sharer} paper~\cite{Carlini-2019-SS}, which utilizes random words in the vocabulary of GPT-2 tokenizer to fill in the corresponding slots.

Note that the cardinality |$\mathcal {R}$| of the randomness space $\mathcal R$ in our experiment is equal to the length of the fill-in candidate list. For each candidate list, we only insert 40$\%$ elements(with the corresponding format) of the whole list to the training set. Moreover, to explore the effect of repetition of canaries, we put them in a specific order and the first element will emerge 10 times in the training set and each of the following elements will repeat 10 more times than its last element. When calculating exposures, they are all taken into consideration along with those who are not inserted into the training set. 

We use \textit{exposure rate} and \textit{mean exposure} as our DEAs' metrics and elaborate on them in this paragraph. Given a specific model, we consider a canary inserted in the training set as \textit{exposed} when its log-perplexity is lower than the perplexities of any canary of the same type not inserted in the training set. 
Thus, an exposed canary has a higher likelihood than all canaries not inserted. 
Note that whether a canary is exposed or not is estimated empirically because we can't exhaust all possible canaries that are not inserted. 
The proportion of exposed canary is defined as \textit{exposure rate}. 
We then investigate the relationship between the frequency of repetitions and the exposure levels experienced by a specific canary type within a given model. 
Both \textit{mean exposure} and \textit{exposure rate} can be representative metrics to be reported. 

We conduct experiments on four decoder-only GPT-2 models: GPT-2\textsubscript{small}, GPT-2\textsubscript{medium}, GPT-2\textsubscript{large}, GPT-2\textsubscript{xl} as well as T5 models: T5\textsubscript{small}, T5\textsubscript{base}, T5\textsubscript{large}, T5\textsubscript{xl} and FLAN-T5\textsubscript{xl} which take encoder-decoder architectures.  
The canary formats are designed differently to adapt to GPT-2 and T5 model architectures. 
Accordingly, when calculating exposure rate and mean exposure, we calculate the likelihood of the canaries for GPT-2 and the conditional likelihood for T5. Table~\ref{app:canary-format} shows the details of the canary format in our experiment.

\subsection{EIA Details}
For the embedding inversion attack, we follow the implementation of GEIAs~\cite{li-etal-2023-sentence} to train a GPT-2 attacker $\Phi$ for each victim LM $f$ based on the auxiliary data $A_I$.

\textbf{Attacker training}.
To train the attacker decoder model, we follow its language modeling objective via manipulating inputs in the representation level.
For a given sentence $x=$``$w_0 w_1 ... w_{u-1}$'' of length $u$, we use $f(x)$ to denote the embedding of $x$ of victim LM $f$.
First, we apply one trainable fully connected layer to align $f(x)$ to be the same dimension as the attacker model's token representation.
We use $Align(f(x))$ to denote the aligned sentence embedding and $\Phi_{emb}(w_i)$ to denote the representation of token $w_i$ of the attacker model.
Second, We concatenate $Align(f(x))$ to the left side of all tokens' representation to obtain the attacker's input: 
\text{
\footnotesize{
$[Align(f(x)), \Phi_{emb}(w_0),\Phi_{emb}(w_1),...,\Phi_{emb}(w_{u-1})]$}
}.
Lastly, we can feed the input before the attacker's first transformer block and apply language modeling objective with the target sequence [$w_0,w_1,...,w_{u-1},$<eos>] by minimizing the cross-entropy loss at each time step.
Here, the <eos> is the special end of sentence token.

\textbf{Evaluation}.
For inference, the attacker $\Phi$ decodes the first token from $Align(f(x))$.
Then tokens are generated iteratively from previous contexts with the sentence embedding till <eos> is reached.
The token-level micro precision/recall/F1 are reported as EIA's evaluation metrics.

\section{Full Evaluation Results}
\label{app: who_results}

In this section, we list our detailed evaluation results of the three attack methods in Table~\ref{tab:full-mnli} and~\ref{tab:full-qnli}.
The comprehensive evaluation results help further support our empirical findings mentioned in the experimental sessions. 
We report \textit{exposure rate} for DEA.
For MIA, \textit{AUC}, \textit{TPR@0.1\%}, and \textit{TPR@1\%} are reported.
For EIA, micro-level \textit{Precision}, \textit{Recall} and \textit{F1-score} are reported. 
In cases where an attack method is inapplicable to a specific model, we denote this with a ``-'' in the table entry.

For full evaluation results on DEAs, as shown in Figure~\ref{fig:app-full-DEA}, we draw a scatter plot to display the variation of mean exposure with different canary repetitions for four different GPT-2 models and two T5 models on the MNLI dataset. 
We use different markers and colors to represent different tuning methods and whether DP is applied. 
We fit a polynomial of order 2 to data points of any specific tuning method and DP implementation. 
Then, we plot a 95\% confidence interval of those data points. 
We have the following observations: 

1) In the case of fine-tuning, an increase in the number of canary insertions results in a noticeable elevation in mean exposure. 
However, this trend becomes less discernible when applying differential privacy to T5 models, which indicates that T5 models are better protected under DP. Mean exposures without DP generally surpass those with DP. 
Intriguingly, for GPT-2 models, an increase in model size leads to a gradual narrowing of the confidence interval gap between models with and without DP. 

2) For prefix tuning, the mean exposure increases when the number of canary insertions increases, and exposures are larger when DP is not employed. 

3) For prompt tuning, as the number of canary insertions increases, we find that mean exposure is not increasing as it does in fine-tuning and prefix tuning settings. The confidence intervals for both scenarios, with and without DP, largely overlap.

\begin{table*}
\centering
\small
\fontsize{9pt}{9pt}\selectfont
  \begin{tabular}{lclc| c ccc| ccc}
    \toprule
    \multirow{3}{*}{Model} &
    \multirow{3}{*}{DP?} &
    \multirow{3}{*}{Tuning Algo} &
    \multirow{3}{*}{Utility} &
      \multicolumn{4}{c|}{Fine-tune Data Privacy} &
      \multicolumn{3}{c}{Inference Data Privacy} 
       \\

      {} & {} & {} & {} & {DEA} & \multicolumn{3}{c|}{MIA}  &\multicolumn{3}{c}{EIA}   \\
      {} & {} & {} & {} & {ER} & {AUC} & TPR@0.1\% & TPR@1\%  & Pre & Rec & F1 \\
      \midrule

    \multirow{6}{*}{GPT-2\textsubscript{small}} 
    & \multirow{3}{*}{Y}         
    & Finetune 
    & 34.96 & 13.89 & 51.09 & 0.12 & 1.03 & 48.95 & 27.53 & 35.24 \\
    &     & Prompt
    & 35.81 & 5.95 & 50.55 & 0.12 & 1.37 & 47.43 & 28.08 & 3527 \\
    &     & Prefix
    & 36.85 & 9.13 & 50.70 & 0.03 & 0.73 & 50.08 & 28.72 & 36.50\\
    \cmidrule(lr){2-11}    
   & \multirow{3}{*}{N}         
    & Finetune 
    & 39.29 & 51.20 & 52.33 & 0.09 & 1.25 & 47.60 & 27.97 & 35.24\\
    &     & Prompt
    & 37.02 & 6.35 & 50.43 & 0.30 & 1.25 & 45.64 & 27.56 & 34.37\\
    &     & Prefix
    & 34.25 & 9.92 & 50.39 & 0.09 & 0.79 & 50.62 & 29.49 & 37.27\\

    \midrule
    \multirow{6}{*}{GPT-2\textsubscript{medium}} 
    & \multirow{3}{*}{Y}         
    & Finetune 
    & 41.31 & 15.08 & 51.33 & 0.06 & 1.00 & 54.84 & 28.22 & 37.27\\
    &     & Prompt
    & 33.25 & 4.37 & 51.10 & 0.15 & 1.46 & 46.60 & 27.60 & 34.67\\
    &     & Prefix
    & 35.17 & 8.73 & 50.22 & 0.18 & 0.85 & 54.93 & 29.97 & 38.78\\
    \cmidrule(lr){2-11}    
   & \multirow{3}{*}{N}         
    & Finetune 
    & 63.37 & 91.27 & 54.49 & 0.21 & 1.37 & 56.63 & 27.52 & 37.04\\
    &     & Prompt
    & 35.01 & 5.95 & 51.12 & 0.15 & 1.34 & 46.97 & 27.56 & 34.74\\
    &     & Prefix
    & 43.86 & 35.32 & 50.77 & 0.21 & 1.31 & 53.74 & 30.34 & 38.78\\

    \midrule
    \multirow{6}{*}{GPT-2\textsubscript{large}} 
    & \multirow{3}{*}{Y}         
    & Finetune 
    & 46.75 & 29.76 & 52.42 & 0.06 & 1.31 & 61.62 & 31.02 & 41.27\\
    &     & Prompt
    & 33.28 & 5.95 & 51.00 & 0.18 & 0.91 & 58.66 & 32.35 & 41.71\\
    &     & Prefix
    & 35.05 & 12.30 & 50.32 & 0.21 & 0.70 & 61.95 & 32.21 & 42.38\\
    \cmidrule(lr){2-11}    
   & \multirow{3}{*}{N}         
    & Finetune 
    & 76.48 & 98.41 & 61.04 & 0.33 & 2.96 & 61.05 & 31.04 & 41.15\\
    &     & Prompt
    & 31.28 & 6.35 & 50.56 & 0.15 & 0.73 & 57.43 & 31.80 & 40.93\\
    &     & Prefix
    & 52.44 & 68.25 & 50.30 & 0.24 & 0.67 & 60.44 & 32.69 & 42.43\\

    \midrule
    \multirow{6}{*}{GPT-2\textsubscript{xl}} 
    & \multirow{3}{*}{Y}         
    & Finetune 
    & 47.73 & 48.81 & 51.67 & 0.03 & 0.73 & 62.18  & 31.59 & 41.90\\
    &     & Prompt
    & 32.37 & 7.94 & 51.17 & 0.27 & 1.67 & 59.05  & 31.71 & 41.26\\
    &     & Prefix
    & 35.17 & 6.35 & 50.21 & 0.06 & 0.51 & 61.76 & 32.77 & 42.82\\
    \cmidrule(lr){2-11}    
   & \multirow{3}{*}{N}         
    & Finetune 
    & 82.88 & 100.00 & 72.87 & 1.28 & 9.92 & 61.75 & 30.89 & 41.18\\
    &     & Prompt
    & 34.37 & 5.95 & 50.91 & 0.42 & 1.28 & 58.42 & 31.36 & 40.81\\
    &     & Prefix
    & 44.37 & 66.27 & 50.62 & 0.12 & 1.00 & 60.81 & 32.99 & 42.78\\

    \bottomrule
  \end{tabular}
  \vspace{-0.05in}
  \caption{\label{tab:full-mnli}
Evaluation results on the MNLI dataset. DEA, MIA and EIA results are reported in \%. The abbreviation ``ER'' represents ``Exposure Rate''.
}
\vspace{-0.15in}
\end{table*}

\subsection{Evaluation with Varied DP Budgets}

\begin{figure*}
\centering
\setlength{\abovecaptionskip}{-0.0cm} 
\subfigure[DEA on GPT-2\textsubscript{medium} with different $\epsilon$.]{
\begin{minipage}[t]{0.48\textwidth}
\centering
\includegraphics[width=\linewidth]{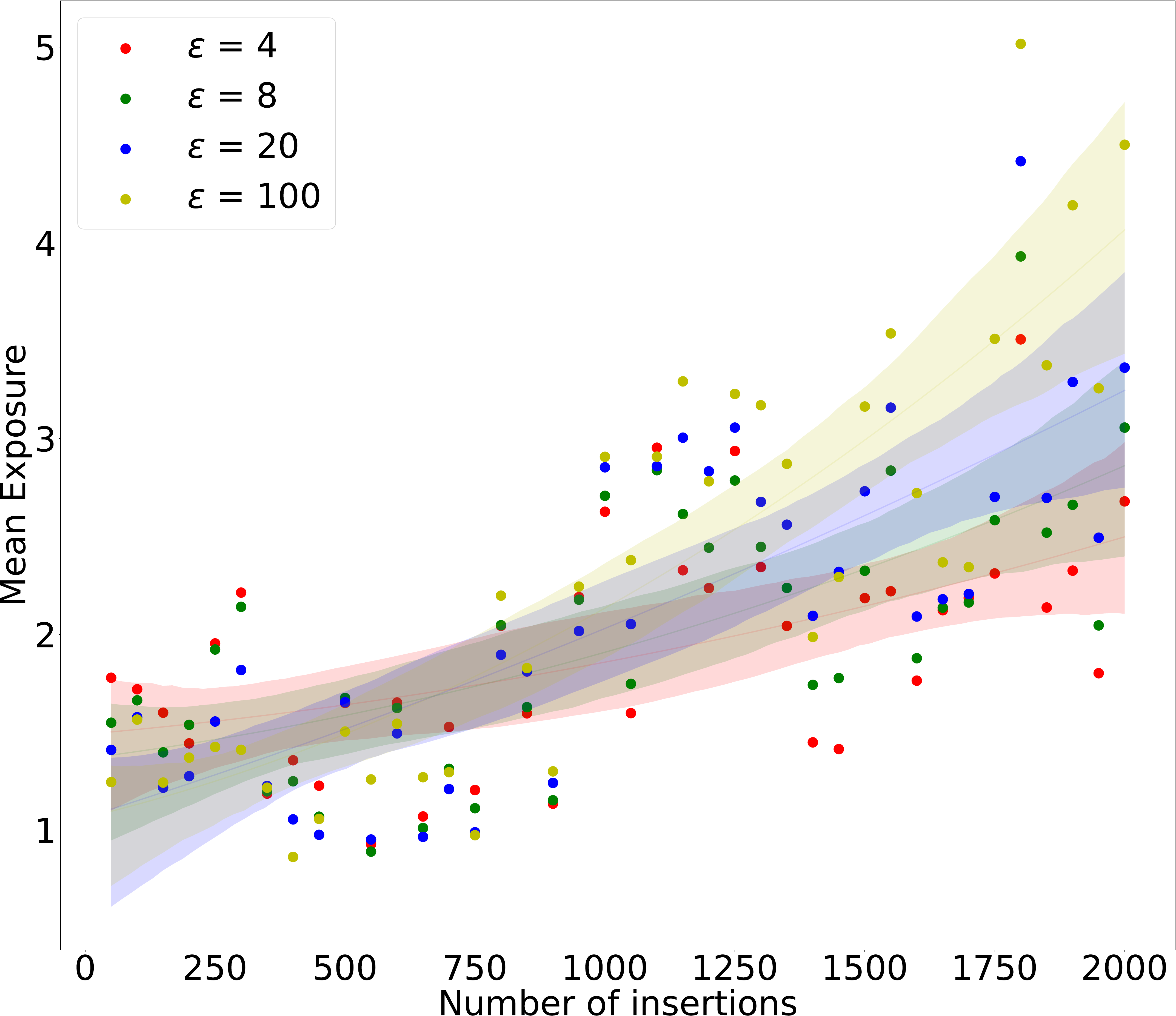}
\end{minipage}
}
\subfigure[DEA on T5\textsubscript{base} with different $\epsilon$.]{
\begin{minipage}[t]{0.48\textwidth}
\centering
\includegraphics[width=\linewidth]{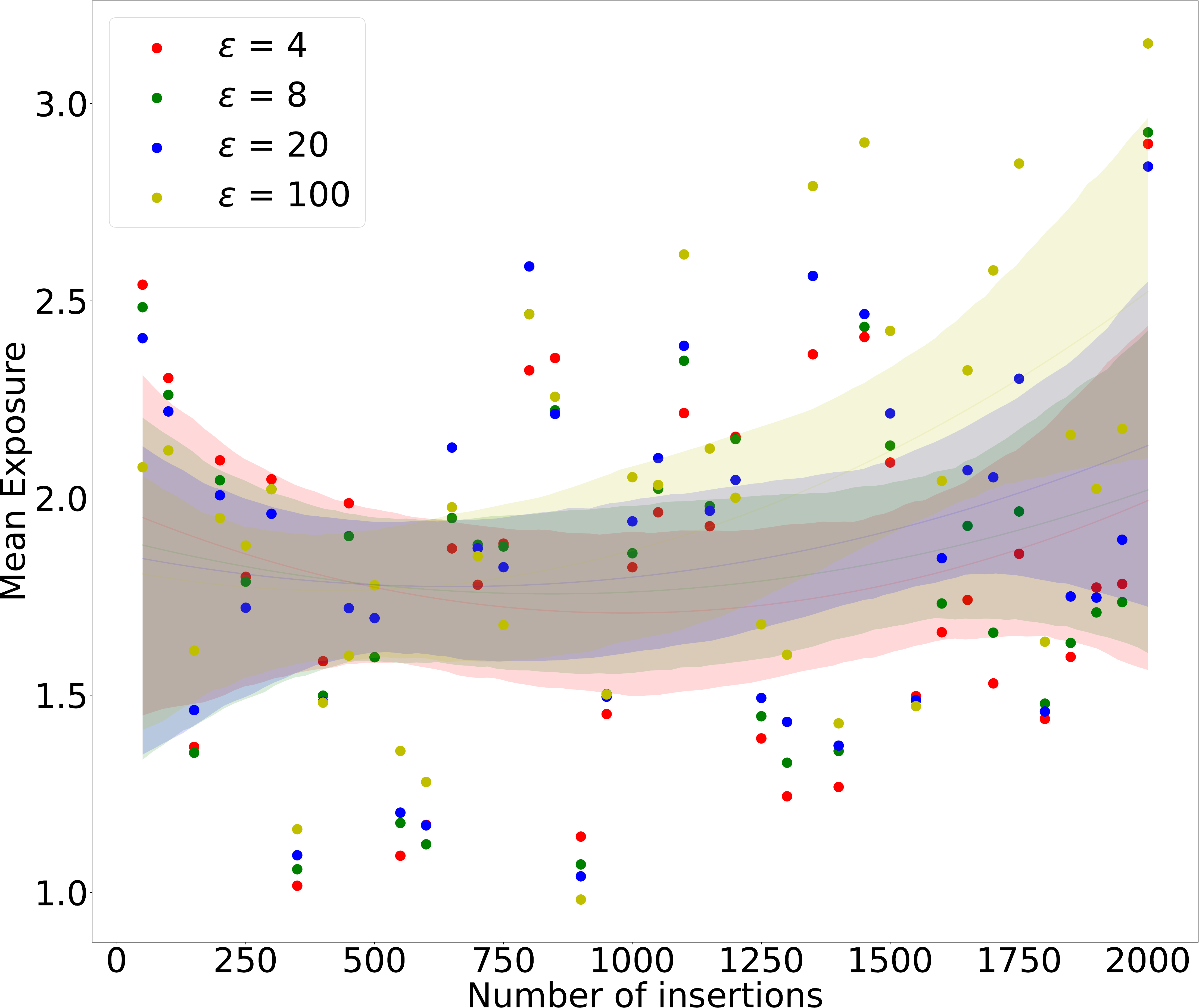}
\end{minipage}
}
\caption{DEA's \textit{mean exposure} evaluation results of GPT-2\textsubscript{medium} and T5\textsubscript{base} with varied $\epsilon$ on the MNLI dataset. 
}\label{fig:different-epsilon}
\vspace{-0.15in}
\end{figure*}

\begin{figure*}[t]
\centering
\setlength{\abovecaptionskip}{-0.0cm} 
\subfigure[DEAs on GPT-2\textsubscript{small}.]{
\begin{minipage}[t]{0.48\textwidth}
\centering\vspace{-0.05in}
\includegraphics[width=\linewidth]{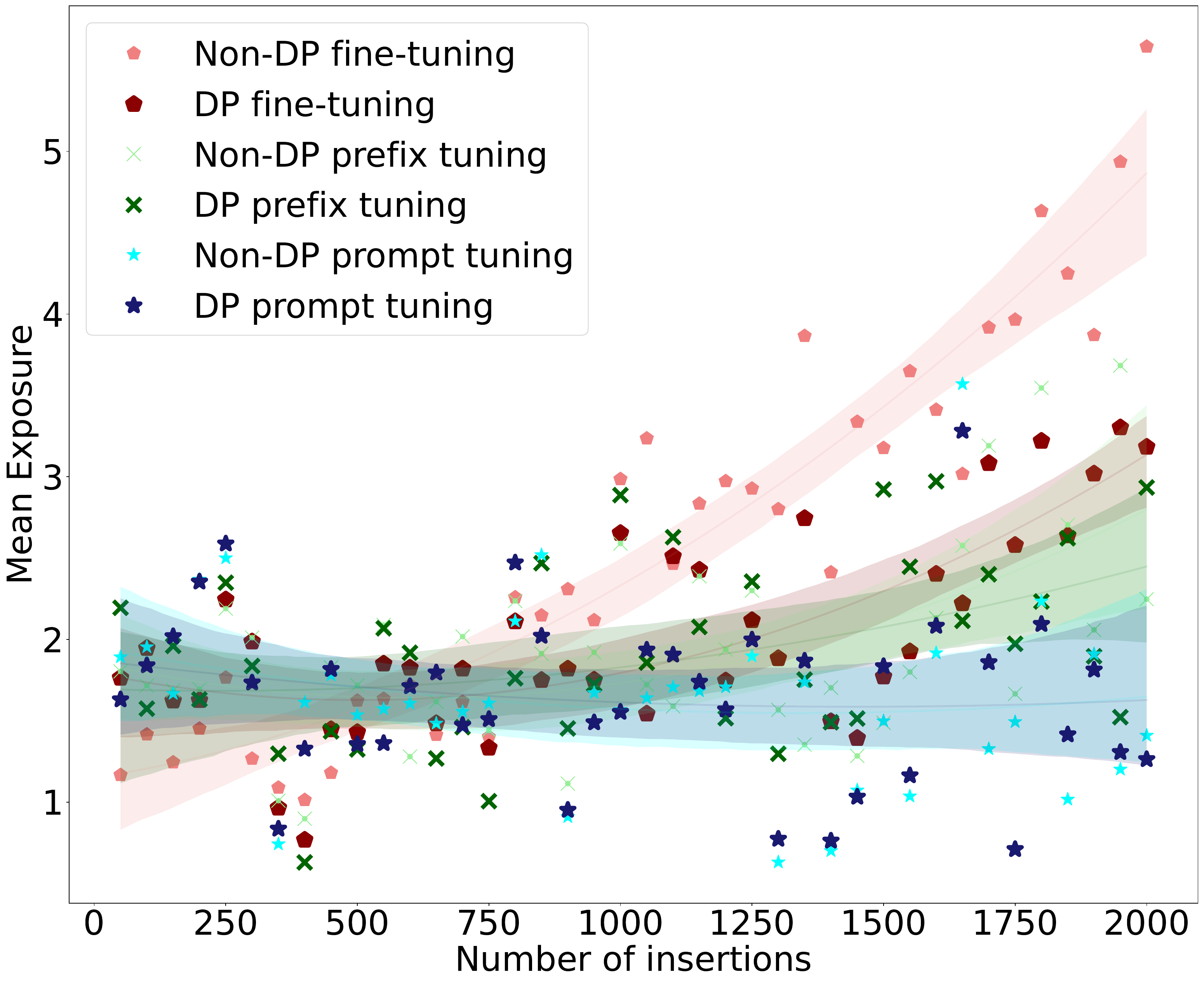}
\end{minipage}
}
\subfigure[DEAs on GPT-2\textsubscript{medium}.]{
\begin{minipage}[t]{0.48\textwidth}
\centering\vspace{-0.05in}
\includegraphics[width=\linewidth]{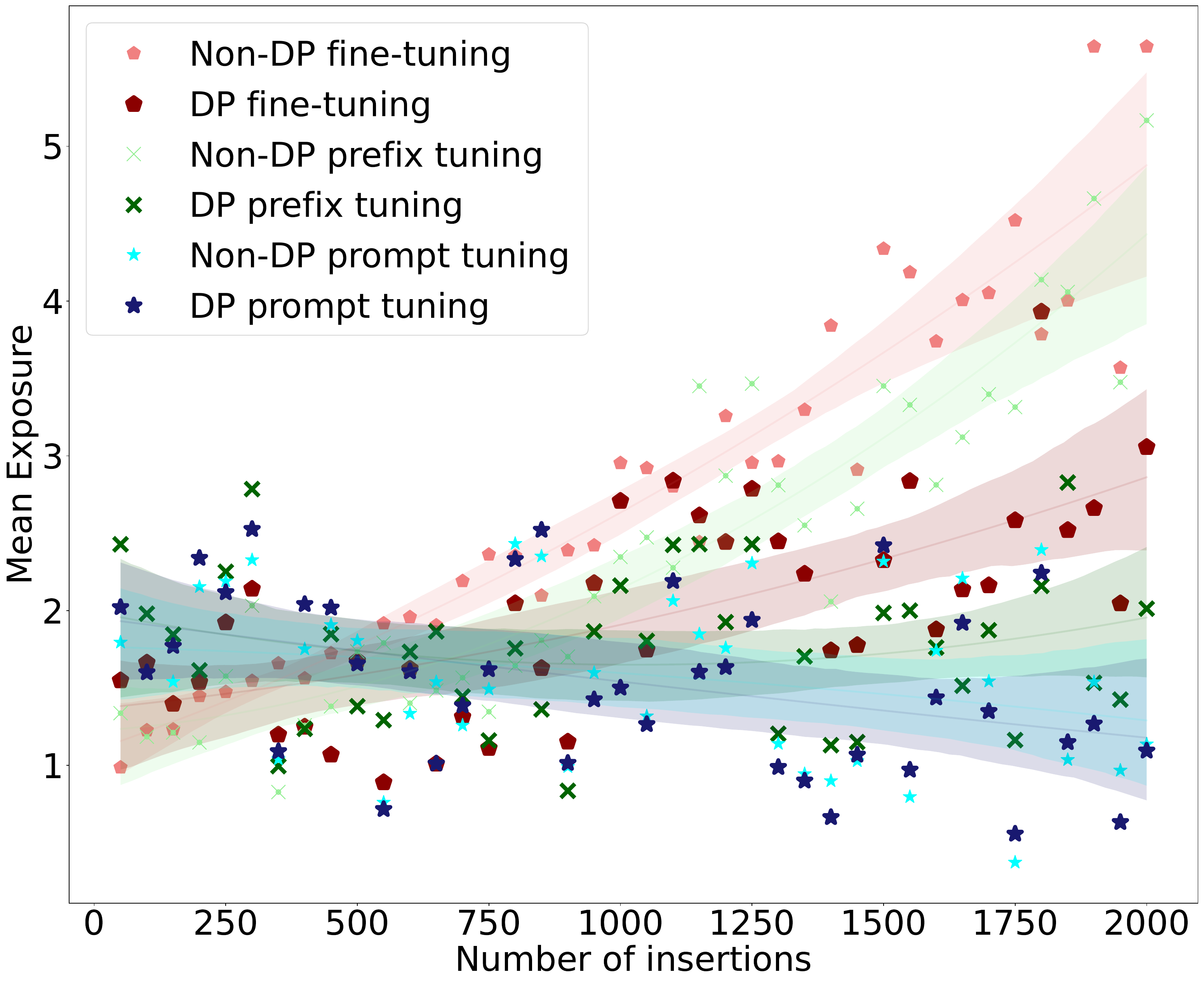}
\end{minipage}
}
\\
\subfigure[DEAs on GPT-2\textsubscript{large}.]{
\begin{minipage}[t]{0.48\textwidth}
\centering\vspace{-0.05in}
\includegraphics[width=\linewidth]{figs/exposure-gpt2-large.pdf}
\end{minipage}
}
\subfigure[DEAs on GPT-2\textsubscript{xl}.]{
\begin{minipage}[t]{0.48\textwidth}
\centering\vspace{-0.05in}
\includegraphics[width=\linewidth]{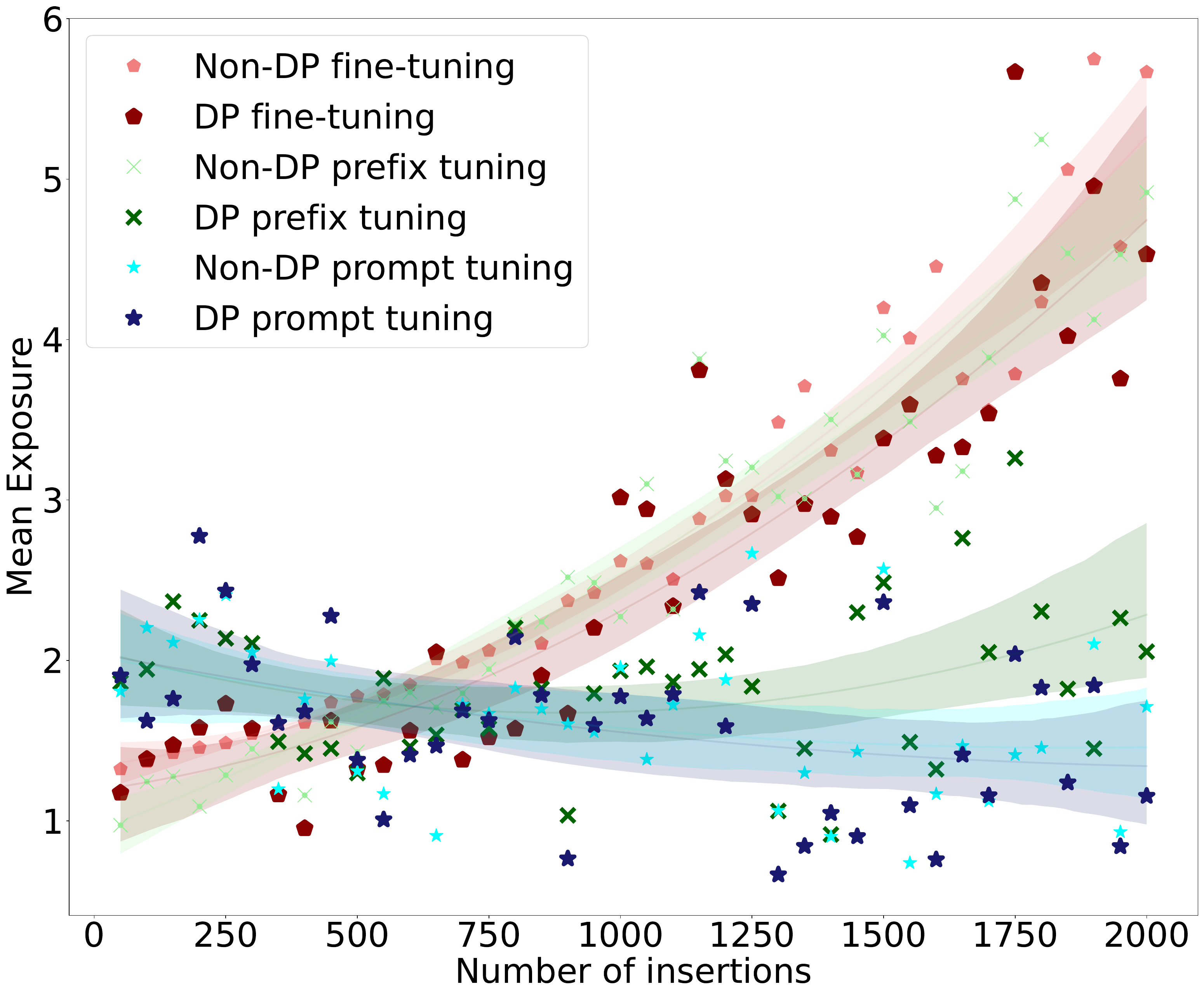}
\end{minipage}
}
\subfigure[DEAs on T5\textsubscript{base}.]{
\begin{minipage}[t]{0.48\textwidth}
\centering\vspace{-0.05in}
\includegraphics[width=\linewidth]{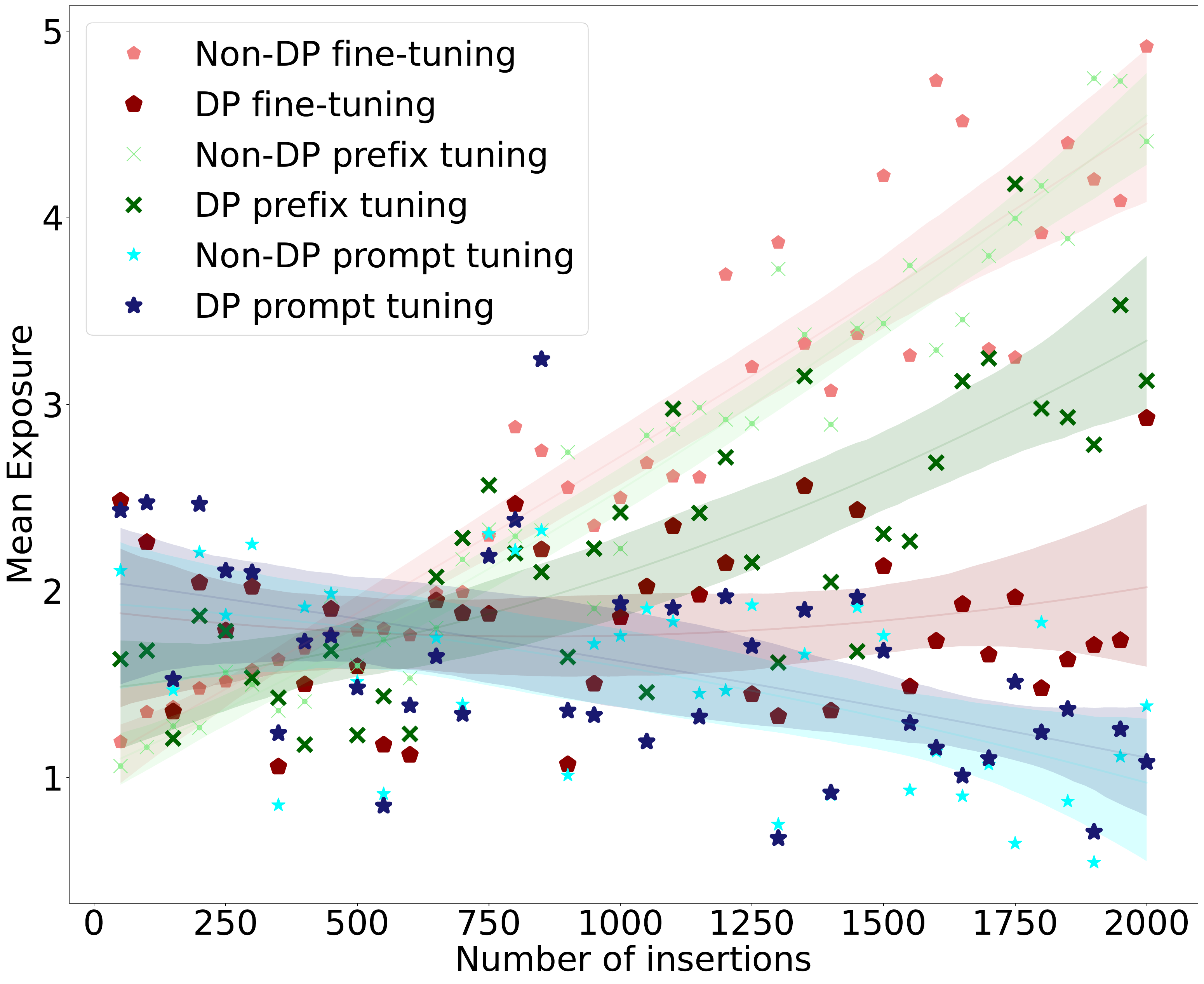}
\end{minipage}
}
\subfigure[DEAs on T5\textsubscript{large}.]{
\begin{minipage}[t]{0.48\textwidth}
\centering\vspace{-0.05in}
\includegraphics[width=\linewidth]{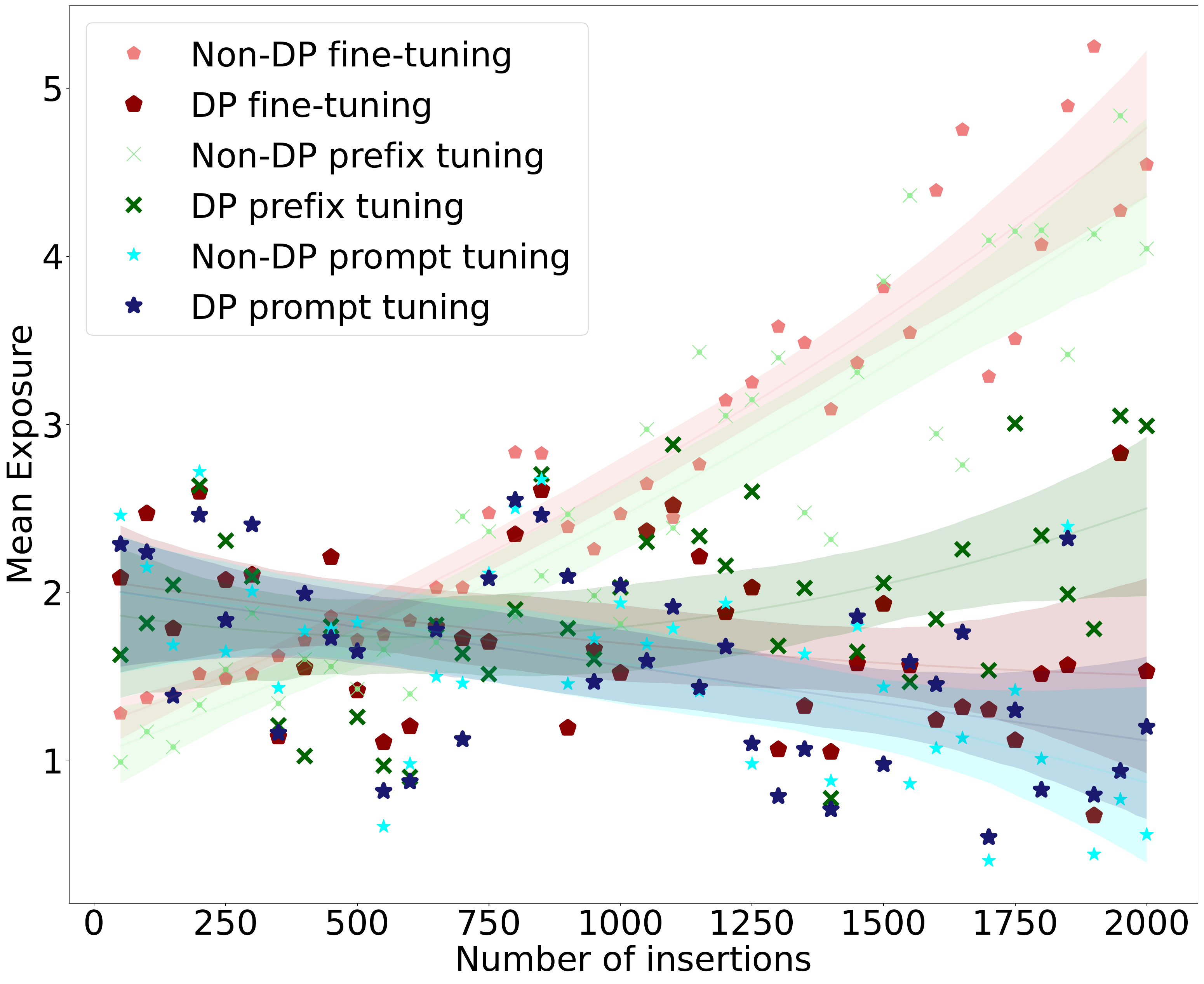}
\end{minipage}
}
\caption{DEA \textit{mean exposure} evaluation results of GPT-2 and T5 models on the MNLI dataset. Note that we obtained the x-axis ``number of insertions'' by multiplying the frequency of canary insertions into the dataset by the total number of training epochs.
}\label{fig:app-full-DEA}
\vspace{-4mm}
\end{figure*}

In this section, we investigated the variation of mean exposure under multiple privacy budget $\epsilon$ on GPT-2\textsubscript{medium} and T5\textsubscript{base}, as shown in Figure~\ref{fig:different-epsilon}. We set $\epsilon$ to be 4, 8, 20, 100 for GPT-2\textsubscript{medium} and T5\textsubscript{base}. 
The results on different $\epsilon$ can be supplementary of the prior results shown in Figure~\ref{fig:app-full-DEA}.

For GPT-2\textsubscript{medium}, the mean exposure increases when the number of canary insertions increases for all $\epsilon$. 
However, for T5\textsubscript{base}, the distribution of mean exposure is more dispersed. 
When the number of canary insertions increases, the increment of mean exposure is not apparent and their confidence intervals largely overlap. 
In addition, the mean exposure of the T5\textsubscript{base} lies in the range of 1.5 to 2.5, which is less than the mean exposure of the GPT-2\textsubscript{medium} that can reach up to 5. 
The above observations suggest that T5 is more robust than GPT-2 under different privacy budget $\epsilon$.

\subsection{DEA Evaluation with Exposure Rate}

\begin{figure*}
\centering
\subfigure[Exposure rate on GPT-2 models.]{
\begin{minipage}[t]{0.48\textwidth}
\centering
\includegraphics[width=\linewidth]{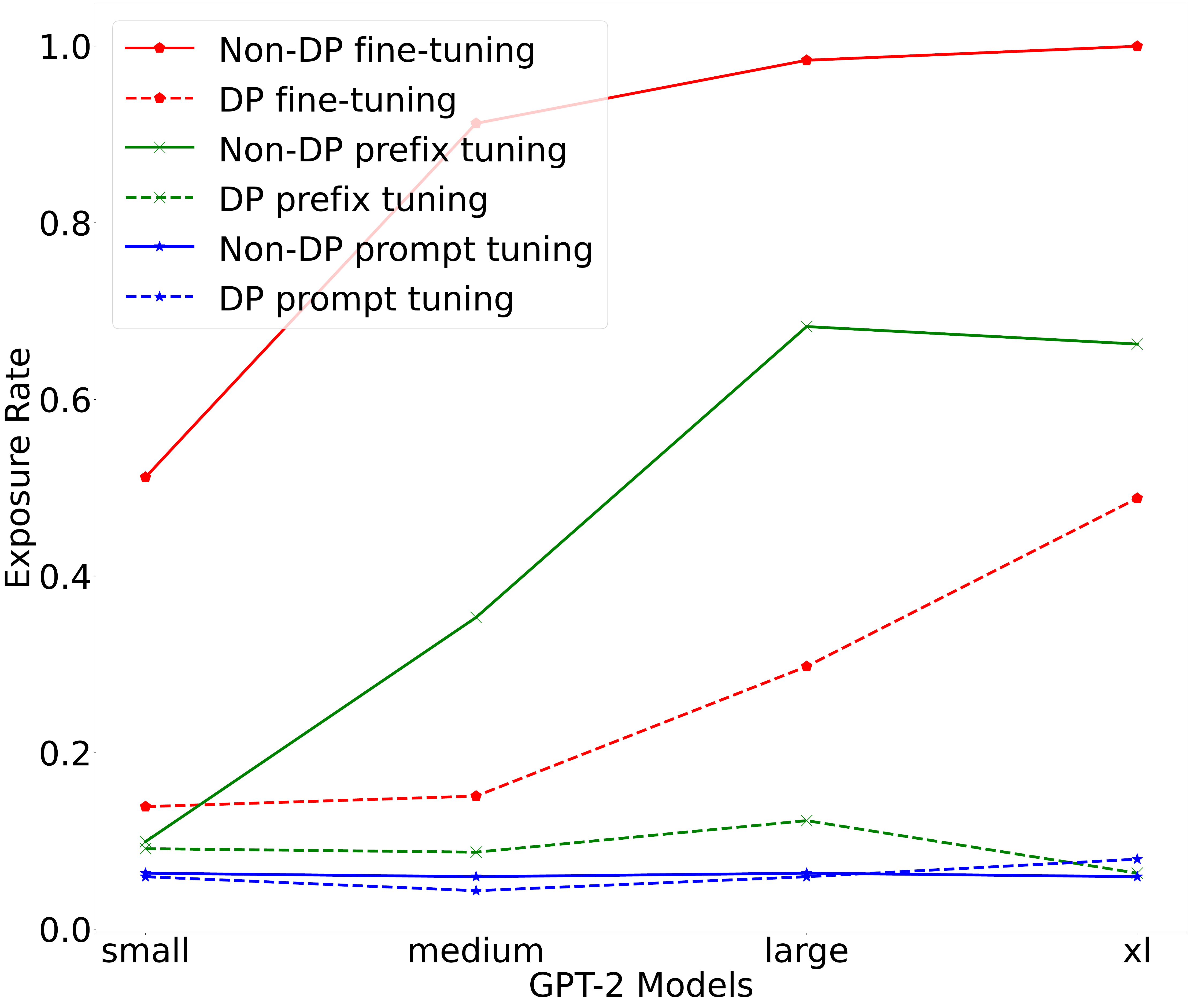}
\end{minipage}
}
\subfigure[Exposure rate on T5 models.]{
\begin{minipage}[t]{0.48\textwidth}
\centering
\includegraphics[width=\linewidth]{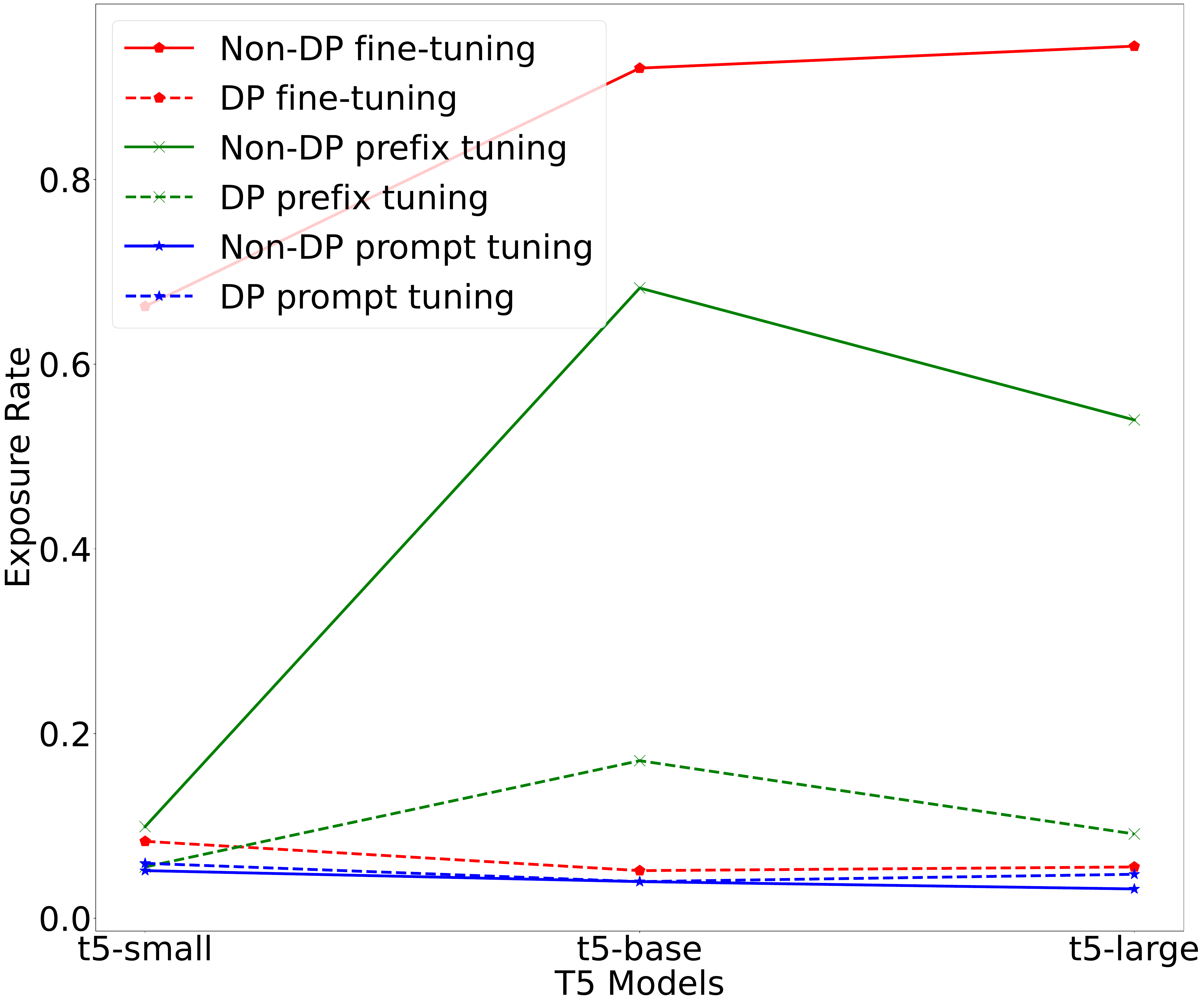}
\end{minipage}
}
\caption{DEA's \textit{exposure rate} evaluation results of GPT-2 and T5 models on the MNLI dataset. 
}\label{fig:exposure-rate}
\vspace{-0.15in}
\end{figure*}

Besides reporting mean exposure, we plot the exposure rate of various GPT-2 and T5 models in Figure~\ref{fig:exposure-rate}. 
We summarize the following observations: 

1) For the fine-tuning setting, the exposure rate gradually increases with the growth of the model size without DP. 
In particular, the exposure rates of T5\textsubscript{large} and \textsubscript{xl} approach 1 in our experiment, which indicate that almost all canaries inserted are exposed. 
We can observe a noticeable decrease in the exposure rate when using DP, which means that DP can be a good protection to avoid sensitive canaries being exposed. 

2) For prefix tuning, as the model size increases, the exposure rate initially rises and then declines for GPT-2 and T5, which suggests that a larger model size does not necessarily increase the risk of the canary being exposed when employing prefix tuning. 

3) For prompt tuning, whether using DP or not, the exposure rate remains consistently low.

\begin{table*}
\centering
\small
\fontsize{9pt}{9pt}\selectfont
  \begin{tabular}{lclc| c ccc| ccc}
    \toprule
    \multirow{3}{*}{Model} &
    \multirow{3}{*}{DP?} &
    \multirow{3}{*}{Tuning Algo} &
    \multirow{3}{*}{Utility} &
      \multicolumn{4}{c|}{Fine-tune Data Privacy} &
      \multicolumn{3}{c}{Inference Data Privacy} 
       \\

      {} & {} & {} & {} & {DEA} & \multicolumn{3}{c|}{MIA}  &\multicolumn{3}{c}{EIA}   \\
      {} & {} & {} & {} & {ER} & {AUC} & TPR@0.1\% & TPR@1\%  & Pre & Rec & F1 \\
      \midrule

      \multirow{8}{*}{BERT\textsubscript{base}} 
    & \multirow{4}{*}{Y}         
    & Finetune 
    & 77.72 & - & 50.12 & 0.11 & 0.98 & 44.22 & 18.05 & 25.63 \\
    &     & Prompt
    & 71.74 & - & 50.06 & 0.25 & 0.94 & 42.44 & 18.20 & 25.47 \\
    &     & Prefix
    & 68.29 & - & 49.96 & 0.12 & 1.08 & 44.62 & 18.41 & 26.06 \\
    &     & Infilling
    & 56.58 & - & - & - & - & 42.23 & 18.12 & 25.36  \\

    \cmidrule(lr){2-11}
    
   & \multirow{4}{*}{N}         
    & Finetune 
    & 90.14 & - & 50.14 & 0.09 & 1.05 & 41.57 & 17.71 & 24.84\\
    &     & Prompt
    & 79.34 & - & 51.40 & 0.15 & 1.36 & 43.80 & 18.53 & 26.05\\
    &     & Prefix
    & 87.45 & - & 51.19 & 0.16 & 1.38 & 44.17 & 18.41 & 25.98\\
    &     & Infilling
    & 85.17 & - & - & - & - & 41.38 & 17.81 & 24.90  \\
    
    \midrule

    \multirow{8}{*}{BERT\textsubscript{large}} 
    & \multirow{4}{*}{Y}         
    & Finetune 
    & 81.14 & - & 49.80 & 0.12 & 0.97 & 42.29 & 17.65 & 24.91 \\
    &     & Prompt
    & 71.55 & - & 50.03 & 0.19 & 1.05 & 40.67 & 17.55 & 24.52 \\
    &     & Prefix
    & 69.05 & - & 49.77 & 0.12 & 1.00 & 42.68 & 17.92 & 25.24 \\
    &     & Infilling
    & 71.47 & - & - & - & - & 40.73 & 17.30 & 24.28\\

    \cmidrule(lr){2-11}
    
   & \multirow{4}{*}{N}         
    & Finetune 
    & 90.93 & - & 50.02 & 0.09 & 1.01 & 41.38 & 17.30 & 24.40 \\
    &     & Prompt
    & 76.07 & - & 49.91 & 0.08 & 0.91 & 41.97 & 17.89 & 25.09 \\
    &     & Prefix
    & 87.94 & - & 50.58 & 0.14 & 1.25 & 43.08 & 17.39 & 24.77 \\
    &     & Infilling
    & 88.89 & - & - & - & - & 40.06 & 17.08 & 23.94\\

    \midrule
    \multirow{6}{*}{RoBERTa\textsubscript{base}} 
    & \multirow{3}{*}{Y}         
    & Finetune 
    & 79.67 & - & 50.28 & 0.07 & 0.92 & 34.57 & 14.35 & 20.28 \\
    &     & Prompt
    & 57.82 & - & 49.83 & 0.17 & 0.90 & 38.87 & 16.07 & 22.74  \\
    &     & Prefix
    & 72.74 & - & 50.03 & 1.01 & 1.01 & 38.72 & 16.25 & 22.89 \\
    \cmidrule(lr){2-11}    
   & \multirow{3}{*}{N}         
    & Finetune 
    & 91.73 & - & 64.90 & 3.34 & 10.39 & 37.77 & 15.81 & 22.29 \\
    &     & Prompt 
    & 80.80 & - & 50.14 & 0.09 & 0.98 & 41.14 & 17.61 & 24.67 \\
    &     & Prefix
    & 89.68 & - & 50.62 & 0.15 & 1.20 & 39.88 & 17.08 & 23.91 \\

    \midrule
    \multirow{6}{*}{RoBERTa\textsubscript{large}} 
    & \multirow{3}{*}{Y}         
    & Finetune 
    & 84.31 & - & 50.20 & 0.11 & 1.02 & 37.96 & 15.62 & 22.13 \\
    &     & Prompt
    & 73.32 & - & 50.55 & 0.15 & 1.16 & 39.47 & 16.40 & 23.17 \\
    &     & Prefix
    & 72.71 & - & 49.74 & 0.62 & 1.06 & 38.12 & 16.15 & 22.69 \\
    \cmidrule(lr){2-11}    
   & \multirow{3}{*}{N}         
    & Finetune 
    & 93.43 & - & 61.12 & 3.33 & 9.25 & 34.17 & 14.66 & 20.52 \\
    &     & Prompt
    & 81.62 & - & 49.83 & 0.08 & 1.01 & 39.45 & 16.40 & 23.17 \\
    &     & Prefix
    & 92.21 & - & 50.39 & 0.08 & 1.11 & 37.76 & 15.98 & 22.45 \\

    \midrule
    \multirow{6}{*}{GPT-2\textsubscript{small}} 
    & \multirow{3}{*}{Y}         
    & Finetune 
    & 50.04 & 11.51 & 52.86 & 0.21 & 1.43  & 50.04 & 19.62 & 28.19 \\
    &     & Prompt
    & 49.77 & 6.75 & 50.15 & 0.15 & 0.82 & 48.02 & 20.56 & 28.79 \\
    &     & Prefix
    & 53.46 & 6.35 & 49.12 & 0.18 & 0.97 & 51.22 & 20.49 & 29.27 \\
    \cmidrule(lr){2-11}    
   & \multirow{3}{*}{N}         
    & Finetune 
    & 52.24 & 38.10 & 55.22 & 0.15 & 2.22 & 53.26 & 20.65 & 29.76 \\
    &     & Prompt
    & 49.77 & 2.78 & 50.96 & 0.09 & 0.81 & 49.22 & 20.28 & 28.72 \\
    &     & Prefix
    & 49.58 & 8.33 & 49.45 & 0.12 & 0.94 & 50.32 & 20.28 & 28.91 \\

    \midrule
    \multirow{6}{*}{GPT-2\textsubscript{medium}} 
    & \multirow{3}{*}{Y}         
    & Finetune 
    & 53.61 & 17.46 & 53.70 & 0.42 & 1.28 & 52.87 & 20.25 & 29.29 \\
    &     & Prompt
    & 51.84 & 5.16 & 50.02 & 0.09 & 0.82 & 52.85 & 20.53 & 29.57 \\
    &     & Prefix
    & 48.51 & 7.54 & 49.86 & 0.12 & 1.06 & 55.84 & 21.50 & 31.04 \\
    \cmidrule(lr){2-11}    
   & \multirow{3}{*}{N}         
    & Finetune 
    & 58.22 & 78.17 & 59.28 & 0.67 & 2.31  & 56.64 & 21.80 & 31.48 \\
    &     & Prompt
    & 52.63 & 4.76 & 50.02 & 0.18 & 0.76  & 53.96 & 20.77 & 29.99 \\
    &     & Prefix
    & 51.02 & 10.31 & 49.87 & 0.27 & 1.15  & 51.08 & 20.47 & 29.22 \\

    \midrule
    \multirow{6}{*}{GPT-2\textsubscript{large}} 
    & \multirow{3}{*}{Y}         
    & Finetune 
    & 55.32 & 31.75 & 52.24 & 0.03 & 1.03 & 58.33 & 22.76 & 32.75 \\
    &     & Prompt
    & 47.29 & 7.94 & 48.64 & 0.18 & 0.97 & 57.96 & 22.58 & 32.50 \\
    &     & Prefix
    & 50.71 & 9.52 & 48.93 & 0.09 & 0.82  & 58.69 & 22.93 & 32.98 \\
    \cmidrule(lr){2-11}    
   & \multirow{3}{*}{N}         
    & Finetune 
    & 67.43 & 97.62 & 67.45 & 0.91 & 4.88 & 59.40 & 22.84 & 32.99 \\
    &     & Prompt
    & 50.19 & 5.95 & 49.69 & 0.09 & 0.82 & 57.65 & 22.62 & 32.49 \\
    &     & Prefix
    & 51.44 & 38.10 & 49.72 & 0.07 & 0.92 & 60.17 & 23.11 & 33.40 \\

    \midrule
    \multirow{6}{*}{GPT-2\textsubscript{xl}} 
    & \multirow{3}{*}{Y}         
    & Finetune 
    & 55.59 & 42.46 & 53.26 & 0.24 & 1.46 & 59.55 & 23.09 & 33.28 \\
    &     & Prompt
    & 50.93 & 7.14 & 49.28 & 0.03 & 0.54 & 58.24 & 22.78 & 32.75 \\
    &     & Prefix
    & 49.67 & 6.75 & 49.52 & 0.09 & 0.85 & 51.08 & 20.47 & 29.22 \\
    \cmidrule(lr){2-11}    
   & \multirow{3}{*}{N}         
    & Finetune 
    & 83.03 & 99.60 & 82.35 & 8.02 & 22.71 & 59.13 & 22.80 & 32.91 \\
    &     & Prompt
    & 51.93 & 5.95 & 48.73 & 0.06 & 0.88 & 58.40 & 22.87 & 32.87 \\
    &     & Prefix
    & 52.33 & 33.73 & 49.88 & 0.18 & 1.00 & 59.02 & 23.15 & 33.25 \\

    \bottomrule
  \end{tabular}
  \vspace{-0.05in}
  \caption{\label{tab:full-qnli}
Full evaluation results on the QNLI dataset.  DEA, MIA and EIA results are reported in \%. The abbreviation ``ER'' represents ``Exposure Rate''.
}
\vspace{-0.05in}
\end{table*}

\end{document}